%% file: arxiv.tex
\documentclass[10pt,twocolumn,letterpaper]{article}

\usepackage{cvpr}
\usepackage{times}
\usepackage{epsfig}
\usepackage{graphicx}
\usepackage{amsmath}
\usepackage{amssymb}
\usepackage{subcaption}
\usepackage{etoolbox}
\usepackage{algorithm,algpseudocode}
\usepackage{multirow}
\usepackage{sidecap}
\usepackage[pagebackref=true,breaklinks=true,letterpaper=true,colorlinks,bookmarks=false]{hyperref}

\newcommand{\MyMapTemplatePrefixc}[4]{\expandafter#1\csname#3#4\endcsname{#2{#4}}} 
\forcsvlist{\MyMapTemplatePrefixc {\def} {\mathcal}{c}} {A,B,C,D,E,F,G,H,I,J,K,L,M,N,O,P,Q,R,S,T,U,V,W,X,Y,Z}  

\newcommand{\MyMapTemplatePrefixtb}[5]{\expandafter#1\csname#4#5\endcsname{#2{#3{#5}}}} 
\forcsvlist{\MyMapTemplatePrefixtb {\def} {\tilde}{\mathbf}{t}} {A,B,C,D,E,F,G,H,I,J,K,L,M,N,O,P,Q,R,S,T,U,V,W,X,Y,Z}  
\forcsvlist{\MyMapTemplatePrefixtb {\def} {\tilde}{\mathbf}{t}} {0,1,a,b,c,d,e,f,g,h,i,j,k,l,m,n,o,p,q,r,s,u,v,w,x,y,z}  

\DeclareMathOperator*{\argmin}{arg\,min}

\newcommand{\MyMapTemplateNoPrefix}[3]{\expandafter#1\csname#3\endcsname{#2{#3}}}
\forcsvlist{\MyMapTemplateNoPrefix {\def} {\mathbf} } {0,1,a,b,c,d,e, f, g, h, i, j, k, l, m, n, o, p, q, r, u, v, w, x, y, z} 
\forcsvlist{\MyMapTemplateNoPrefix {\def} {\mathbf} } {A,B,C,D,E,F,G,H,I,J,K,L,M,N,O,P,Q,R,S,T,U,V,W,X,Y,Z}  

\def\ie{{\it i.e., }}

\usepackage{color,colortbl}
\definecolor{mypink1}{rgb}{0.858, 0.188, 0.478}
\definecolor{mypink2}{RGB}{219, 48, 122}
\definecolor{mypink3}{cmyk}{0, 0.7808, 0.4429, 0.1412}
\definecolor{mygray}{gray}{0.6}
\definecolor{myblue}{rgb}{0.2, 0.3, 0.8}

\cvprfinalcopy 

\ifcvprfinal\fi
\begin{document}

\title{Cluster, Split, Fuse, and Update: Meta-Learning for Open Compound Domain Adaptive Semantic Segmentation}

\author{Rui Gong \textsuperscript{\rm 1}, Yuhua Chen \textsuperscript{\rm 1}, Danda Pani Paudel \textsuperscript{\rm 1}, Yawei Li \textsuperscript{\rm 1}, Ajad Chhatkuli \textsuperscript{\rm 1}, \\ Wen Li \textsuperscript{\rm 3}, Dengxin Dai \textsuperscript{\rm 1}, Luc Van Gool \textsuperscript{\rm 1,2}\\
\textsuperscript{\rm 1} Computer Vision Lab, ETH Zurich, \textsuperscript{\rm 2} VISICS, KU Leuven, \textsuperscript{\rm 3} UESTC\\
\{gongr, yuhua.chen, paudel, yawei.li, ajad.chhatkuli, dai, vangool\}@vision.ee.ethz.ch, \\ liwenbnu@gmail.com
}
\maketitle
\thispagestyle{empty}

\input{00_Abstract}
\input{01_Introduction}
\input{02_RelatedWorks}
\input{03_method}
\input{05_Experimetns}
\input{06_Conclusion}

{\small
\bibliographystyle{ieee_fullname}
\bibliography{07_egbib}
}

\clearpage
\input{08_Supplementary}

\end{document}

%% file: 00_Abstract.tex
  \begin{abstract}
    Open compound domain adaptation (OCDA) is a domain adaptation setting, where target domain is modeled as a compound of multiple unknown homogeneous domains, which brings the advantage of improved generalization to unseen domains. In this work, we propose a principled meta-learning based approach to OCDA for semantic segmentation, MOCDA, by modeling the unlabeled target domain continuously.
     Our approach consists of four key steps. First, we \textbf{cluster} target domain into multiple sub-target domains by image styles, extracted in an unsupervised manner. Then, different sub-target domains are \textbf{split} into independent branches, 
     for which batch normalization parameters are learnt to treat them independently.
     A meta-learner is thereafter deployed to learn to \textbf{fuse} sub-target domain-specific predictions, conditioned upon the style code. Meanwhile, we learn to online \textbf{update} the model by model-agnostic meta-learning (MAML) algorithm, thus to further improve generalization. 
     We validate the benefits of our approach by extensive experiments on synthetic-to-real knowledge transfer benchmark datasets, where we achieve the state-of-the-art performance in both compound and open domains.
\end{abstract}

%% file: 01_Introduction.tex
\section{Introduction}
Semantic segmentation with minimal supervision is one of the most sought-after goals of image understanding. Unfortunately, the learned understanding in one domain does not generalize to the images from other domains~\cite{ben2010theory}. In such cases, domain adaptation aims at transferring the shared knowledge across different but related domains~\cite{pan2010domain}, \ie source and target, using the unlabeled images from the target. When the target domain images  are collected in mixed, continually varying, and even unseen conditions, understanding images invites the problem of open compound domain adaptation~\cite{liu2020open}. 

\begin{figure}
    \centering
    \includegraphics[width=1.0\linewidth]{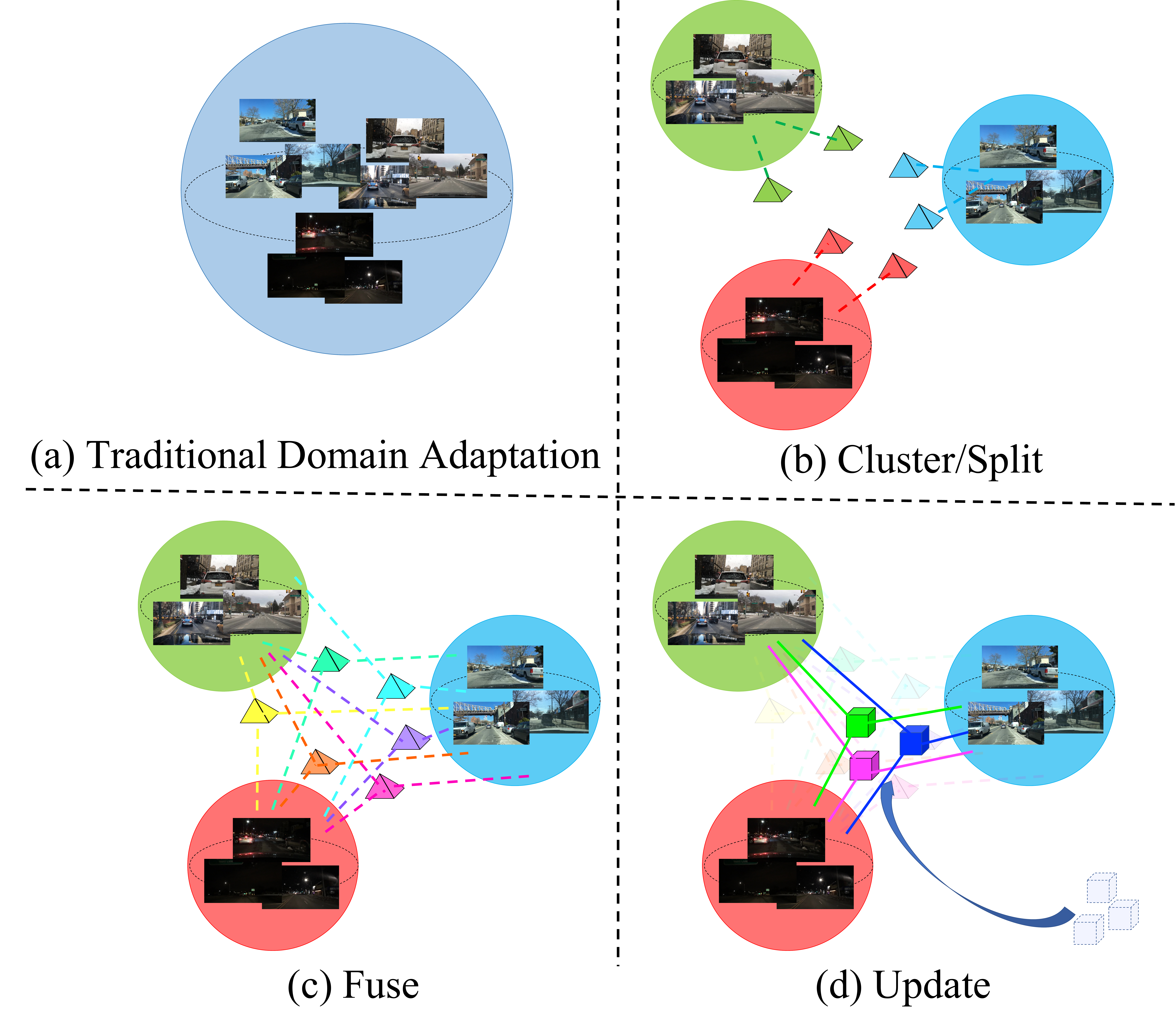}
    \caption{(a) The traditional unsupervised domain adaptation (UDA) vs. (b,c,d) the proposed meta-based open compound domain adaptation (MOCDA). Unlike the traditional UDA,
    MOCDA treats target as a compound of multiple unknown sub-domains. These sub-domains are discovered and processed  using  cluster and split modules (b). The fuse module (c) then linearly combines the sub-domain splits as basis (dash lines).
    On open domains, MOCDA adapts through online update during inference (blue arrow) in (d). Meta-learning serves in the fuse module and the update module.}
    \label{fig:hardsoftlabel}
\end{figure}

The Open Compound Domain Adaptation (OCDA) treats the target as a compound of multiple unknown sub-domains. Such assumption has been shown to be very promising by {\it Liu et al.}~\cite{liu2020open} for many practical settings of image classifications. However, the method developed in \cite{liu2020open} does not fully exploit the same assumption for the task of image segmentation.~\footnote{OCDA~\cite{liu2020open} does not fully exploit the domain information for segmentation task due to the inaccessibility of the domain encoder.Refer the original paper~\cite{liu2020open} for details.}
In this work,  we show that the homogeneous sub-domain assumption can be exploited effectively also for image segmentation. We propose a novel meta-learning based approach to OCDA (abbreviated as MOCDA) that consists of four modules: cluster; split; fuse; and update, as illustrated in Fig.~\ref{fig:hardsoftlabel}.

Similar to OCDA, the proposed MOCDA utilizes two image sets for training from: a single labeled source domain; and a diverse unlabeled target domain, which is assumed to be a compound of multiple unknown sub-domains. Such an assumption is suitable for real challenging situations, where the target domain is a combination of many factors including diverse weather, city, and the acquisition time~\cite{pitropov2020canadian,cordts2016cityscapes,maddern20171}. The considered learning setup not only performs domain adaptation to the compound target domain, but also has generalization potential to unseen open domains. 
In this context, the process of domain adaptation happens to exhibit a meta-behaviour~\cite{li2017learning,balaji2018metareg,chen2019blending}, which learned dynamically makes the open world semantic segmentation possible. In this work, we show that the meta-behaviour of OCDA can be learned using (a) a hypernetwork for dynamic fusion of knowledge, and (b) the online update.  On the one hand, the update process -- which is carried out using the model-agnostic meta-learning strategy  -- creates an opportunity for better open set generalization with  only one gradient step. On the other hand, the learned dynamic fusion allows images to appear from the continuous manifold of the compound target domain.

In essence, the proposed framework serves in following four steps. (i) From target images, style codes are extracted and grouped into multiple clusters.  (ii) For each cluster, a set of batch normalization (BN) parameters are learned.  (iii) Corresponding to each cluster, each image can have different domain-specific predictions. The hypernetwork, then, learns to fuse these predictions. (iv) Model-agnostic meta-learning (MAML)~\cite{finn2017model} is exploited during hypertraining process, endowing the online update ability of the model on open domain during inference stage. The key contributions of this paper can be summarized as follows:
\begin{itemize}
    \item We propose a novel framework for semantic segmentation in the OCDA setting. We use meta-learning in the dynamic fusion and MAML strategy based online update, to address the limitations of~\cite{liu2020open}.
    \item We propose to model the compound target domain continuously, taking the sub-target domain as the basis, which  offers the advantage of adapting to target domain and generalizing to unseen open domains.
    \item We demonstrate the adequacy of image style features, learned in an unsupervised manner, for our meta-based method MOCDA.
    \item The proposed method provides the state-of-the-art results in synthetic-to-real knowledge transfer benchmark datasets, for both compound and open domains.
\end{itemize}

%% file: 02_RelatedWorks.tex
\section{Related Works}
\noindent\textbf{Unsupervised Domain Adaptation and Generalization.}
Our work is related to domain adaptation~\cite{saenko2010adapting,pan2010domain,torralba2011unbiased, gong2012geodesic,venkateswara2017deep,peng2019moment} and domain generalization~\cite{li2017learning,li2017deeper,li2018deep,li2018domain} works. Unsupervised domain adaptation aims at training a model on the labeled source domain and transferring the learned knowledge to the unlabeled target domain. The traditional unsupervised domain adaptation works \cite{long2015learning, ganin2015unsupervised, long2017deep, tzeng2017adversarial} typically focus solving adaptation problem from a single source domain to a single target domain. Even though being effective in several tasks, the single target domain assumption is still restricted in many practical applications. Recently, multiple-target domain adaptation problem~\cite{chen2019blending, gholami2020unsupervised} have received increasing research interests. The problem investigates knowledge transfer to multiple unlabeled target domains. Yet another important aspect not prioritized by the classical domain adaptation methods is the knowledge transfer to unseen but related open domains \cite{liu2020open, gong2019dlow, li2017domain}.

\noindent\textbf{Cross-Domain Semantic Segmentation.}
Cross-domain semantic segmentation is an extensively studied topic, both in the setting of domain adaptation~\cite{zhang2017curriculum,sankaranarayanan2017unsupervised,zou2018unsupervised,chen2018road,chen2019learning,vu2019advent} and in the setting of domain generalization~\cite{volpi2018generalizing,dou2019domain,gong2019dlow,qiao2020learning,liu2020open}. Most works either assume the target domain as a single domain~\cite{sankaranarayanan2017unsupervised,zhang2017curriculum,saito2018maximum,hoffman2018cycada,chen2018road,chen2019learning,zou2018unsupervised}, or a composition of multiple known domains~\cite{gong2019dlow,zhang2019category,zhao2019multi,qiao2020learning}, with an exception of OCDA~\cite{liu2020open}. OCDA assumes target domain as a composition of multiple unknown domains, which is more realistic in practice. \cite{liu2020open} follows a different approach for semantic segmentation compared to the classification task. The curriculum learning therefore is based on the average class confidence scores, rather than the neatly learned domain-focused factors in case of the classification task.    
Nevertheless, the experimental setup of our work is inspired by~\cite{liu2020open}.

\noindent\textbf{Meta-Learning for Domain Adaptation/Generalization.}
Meta-learning addresses the problem of learning to learn and has been successfully applied to various applications including image classification~\cite{ha2016hypernetworks}, image restoration~\cite{hu2019meta}, visual tracking~\cite{bhat2019learning}, and network compression~\cite{li2020dhp}.
The principle of meta-learning~\cite{schmidhuber1987evolutionary,hochreiter2001learning} has also been investigated for the task domain adaptation~\cite{qian2019domain,li2020online,chen2019blending} and generalization~\cite{li2017learning,balaji2018metareg, dou2019domain}, with the algorithmic advances~\cite{andrychowicz2016learning,finn2017model,rajeswaran2019meta}.
Our work can be related to those works in terms of general methodology. Among those works, the ones most related are \cite{chen2019blending} and \cite{zhang2020generalizable}. The similarities are : 1) both of \cite{chen2019blending} and our MOCDA study the domain adaptation problem when there are multiple unknown target domains through meta-learning. 2) both of \cite{zhang2020generalizable} and our MOCDA aims at improving the domain generalization performance for semantic segmentation model, with the help of MAML strategy. However, we have significant differences in the following aspects: 1) \cite{chen2019blending} utilizes the meta-learner for clustering the target domain into different sub-target domains, and the target domain is modeled as a union of multiple sub-target domains. And \cite{chen2019blending} does not include the open domain. However, our meta-hypernetwork is utilized to fuse the knowledge from different clusters, to model the target domain as a continuous compound target domain. 2) \cite{zhang2020generalizable} does not study the domain adaptation problem, and only focus on the domain generalization. The MAML strategy in \cite{zhang2020generalizable} is only used during training stage on the well labeled source domain. By contrast, MOCDA utilizes the MAML strategy in both of the well labeled source domain and the unlabeled target domain during the training stage. During inference, the MAML strategy is exploited to online update our model. 

%% file: 03_Method.tex
\section{The MOCDA Model}
\begin{figure*}
    \centering
    \includegraphics[width=\linewidth]{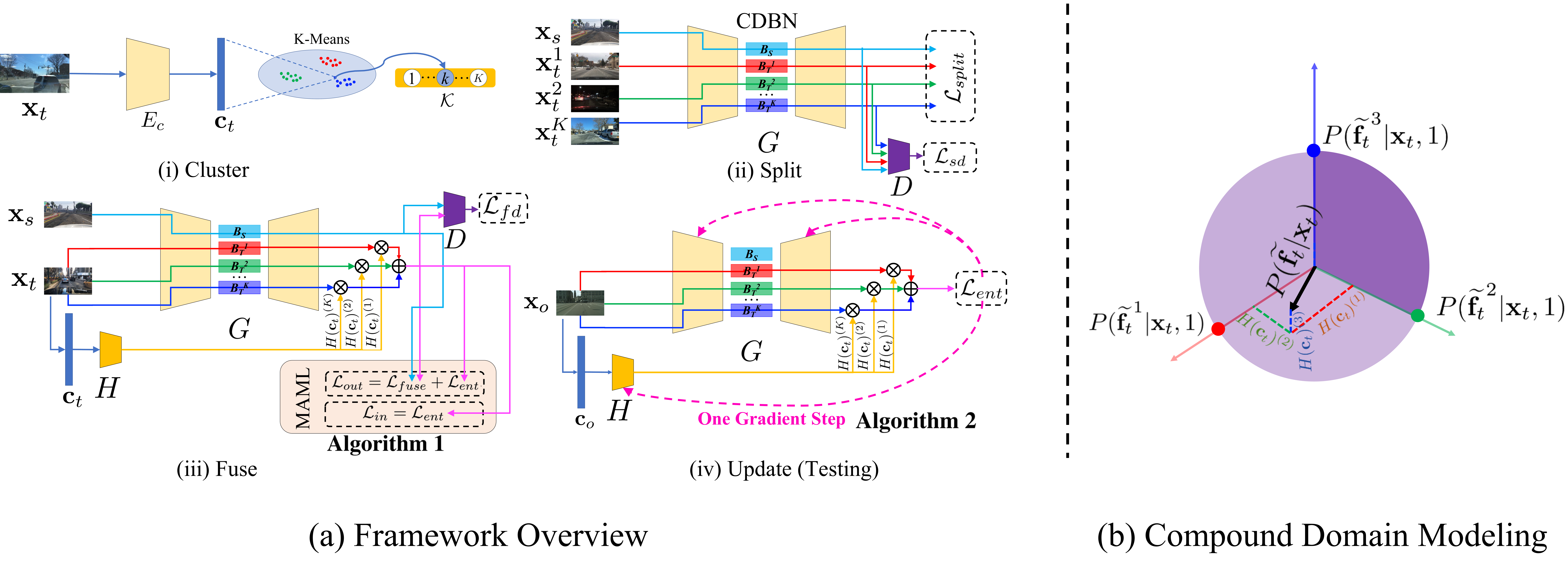}
    \caption{(a) The overview of MOCDA framework demonstrating four modules; (i) Cluster, (ii) Split, (iii) Fuse, and (iv) Update. 
    (b) Illustration of compound domain modeling, taking $\textstyle K=3$ for example. The sub-target domain $\textstyle P(\widetilde{\f_{t}}^{1}|\x_{t}, 1)$, $\textstyle P(\widetilde{\f_{t}}^{2}|\x_{t}, 2)$ and $\textstyle P(\widetilde{\f_{t}}^{3}|\x_{t}, 3)$ is taken as basis. The cluster/split module models the compound target domain as the union set of three points, \ie red, green and blue points. But the fuse module models the compound target domain $\textstyle P(\widetilde{\f_{t}}|\x_{t})$ as the vector $\textstyle H(\c_{t})=[H(\c_{t})^{(1)}, H(\c_{t})^{(2)}, H(\c_{t})^{(3)}]'$, composing the purple half quarter-spherical surface.}
    \label{fig:framestruct}
\end{figure*}
\textbf{Preliminaries.}
We consider that the labeled source domain $\cS$ is composed of the source images $\x_{s}$, and the corresponding semantic labels $\y_{s}$, \ie $\cS = \{(\x_{s}, \y_{s}) | \x_{s} \in \mathbb{R}^{H\times W\times 3}, \y_{s} \in \mathbb{R}^{H\times W}\} $, where $H, W$ are height and width of the image, respectively. In OCDA, the unlabeled target domain $\cT$ consists of target images $\x_{t}^{i}$  from multiple homogeneous sub-target domains, $\cT^{i} = \{ \x_{t}^{i} | \x_{t}^{i} \in \mathbb{R}^{H\times W\times 3}\}, i = 1, \dots N$, where $N$ is number of sub-target domains. In the context of this work (and also in OCDA), these sub-target domains are unknown. Therefore, the images $\x_{t}^{i}$ from some unknown sub-target domain $\cT^{i}$ are simply denoted as $\x_{t}$, for notation convenience and clarity.

In this section, we propose the MOCDA model for semantic segmentation. The MOCDA model is composed of four modules: cluster, split, fuse, and update. The \textbf{Cluster} module extracts and clusters the style code from the target domain images automatically, dividing the target domain into multiple sub-target domains. The \textbf{Split} module adopts the compound-domain specific batch normalization (CDBN) layer to process different sub-target domain images using different branches. The \textbf{Fuse} module exploits a hypernetwork to predict the weights corresponding to each branch adaptively, conditioned on the style code of the input image. The final output of the network is the weighted combination of the outputs of different branches. The MAML method is utilized to train the \textbf{Fuse} module, so as to make the model be adapted quickly in \textbf{Update} module. Finally, the \textbf{Update} is carried out online during the inference time with one-gradient step, which is found to be beneficial for open domains. The framework overview is shown in Fig.~\ref{fig:framestruct}\textcolor{red}{a}. In the following, we provide the details of all four modules, separately.

\subsection{Cluster: Style Code Extraction and Clustering}
The aim of the cluster module is to cluster the target domain $\cT$ into different sub-target domains $\cT^{k}, k=1,\ldots, K$, serving the OCDA's assumptions of unknown multiple sub-target domains of the target domain. As shown in \cite{liu2020open, huang2018multimodal}, the major differences of the target domain images due to varying conditions, such as the weather, lighting, and inter-dataset, can be effectively reflected by the style of the images. Our cluster module consists of two mappings; $E_{c}(\cdot)$ and $E_{l}(\cdot)$. $E_{c}(\cdot)$ maps the target domain $\cT$ to the style code domain $\cC_{t} = \{\c_{t}|\c_{t}\in \mathbb{R}^{l}\}$ as $E_{c}: \cT\rightarrow \cC_{t}$, where $l$ is the dimension of the style code. More specifically, the target domain image $\x_{t}$ is mapped to a low-dimension style code $\c_{t} = E_{c}(\x_{t})$. Then a clustering algorithm, K-means~\cite{lloyd1982least}, is adopted to automatically cluster the style code domain $\cC_{t}$, partitioning into $K$ clusters with centroids \{$\c_{t}^{k}\}$. We use the mapping $E_{l}(\cdot)$ to assign $\x_{t}$ to  one of the sub-target domains, represented by the set ${\cK=\{k|k = 1,\ldots, K\}}$, as $E_{l}: \cT\rightarrow \cK$. Here, we adopt the nearest neighbor strategy for $E_{l}(\cdot)$. More specifically, each target image is assigned to the nearest cluster, using the Euclidean distance between style codes of the image and the centroids, given by,
\begin{equation}\label{eq:nearestNeighbour}
    E_{l}(\x_{t}) :=  \argmin_{k} \|\c_{t}-\c_{t}^{k}\|.
\end{equation}

The key of our cluster module is to find an adequate mapping $E_{c}(\cdot)$. In this work, the unsupervised image translation framework MUNIT \cite{huang2018multimodal} is trained to translate between the source domain $\cS$ and the target domain $\cT$. During the translation training process, the style code encoder of MUNIT is trained to extract the style code from images unsupervisedly. The trained style encoder of MUNIT is used as $E_{c}(\cdot)$. Then, the target domain $\cT$ is clustered into $K$ sub-target domains $\cT^{k}$, where the number of sub-target domains $K$ is a hyperparameter. Using the nearest neighbour search, refer Eq.~\eqref{eq:nearestNeighbour}, each target image $\x_{t}$ is assigned to one of the sub-target domains $\cT^{k}$. Henceforth, the image $\x_{t}$ assigned image $k^{\text{th}}$ cluster is denoted as $\x_{t}^{k}$.

\subsection{Split: Domain-Specific Batch Normalization} \label{sec:split}
In~\cite{Chang_2019_CVPR}, the domain-specific batch normalization (DSBN) is shown to be beneficial for the unsupervised domain adaptation (UDA), by separating the batch normalization layer for the source and target domain.

Similar to DSBN for UDA, the aim of our split module is to separate the multiple sub-target domain-specific information from the domain-invariant information. We propose DSBN for OCDA (abbreviated as CDBN), to conduct such separation for source domain $\cS$ and the multiple (clustered) sub-target domains $\{\cT^{k}\}$.
Note that DSBN for UDA learns only two sets of BN parameters (with possible extension given more labeled domains). However, the proposed CDBN learns $K+1$ sets of BN parameters for source domain and multiple \emph{unlabeled} sub-target domains, \ie $B_{S}, B_{T}^{1},..., B_{T}^{K}$, formulated as,
\begin{eqnarray}
    B_{S}(\x_{s}, \mu_{s}, \sigma_{s}, \beta_{s}, \gamma_{s})
    = \gamma_{s} \frac{\x_{s} - \mu_{s}}{\sigma_{s}} + \beta_{s}, \\
    B_{T}^{k}(\x_{t}^{k}, \mu_{t}^{k}, \sigma_{t}^{k}, \beta_{t}^{k}, \gamma_{t}^{k})
    = \gamma_{t}^{k} \frac{\x_{t}^{k} - \mu_{t}^{k}}{\sigma_{t}^{k}} + \beta_{t}^{k},
\end{eqnarray}
where $k$ is the sub-target domain label, $k = 1, ..., K$.
Our split module replaces BN layers by CDBN. As shown in Fig.~\ref{fig:framestruct}\textcolor{red}{a}, our split module includes the multi-branch semantic segmentation network $G=\{G_{s}, G_{1}, ..., G_{K}\}$ and the discriminator $D$. $G_{k}$ is formed by selecting the k-th branch $B_{k}$ of the CDBN layer. 
Through the adversarial learning, the discriminator $D$ aligns the prediction distributions of source domain and that of the sub-target domains, in the output space. Therefore, the full optimization objective of the split module includes the semantic segmentation loss and the adversarial loss, presented below.

\textbf{Semantic Segmentation Loss.}  We train the semantic segmentation network $G$ with a standard cross entropy loss, using the source domain image $\x_{s}$ and the associated ground truth label $\y_{s}$, 
\begin{eqnarray}
    \cL_{seg}(G) = -\frac{1}{HW}\sum_{n=1}^{HW}\sum_{m=1}^{M}y_{s}^{(n,m)}\log(G_{s}(\x_{s})^{(n,m)}),
    \label{eq:seg_loss}
\end{eqnarray}
where $(n,m)$ represents (pixel, class) indices for $M$ classes.

\textbf{Multi-Branch Adversarial Loss.}  Recall the cluster module, each target image $\x_{t}$ is assigned to a unique sub-target domain label $k$, \ie $\x_{t}^k$. Here in the split module, the image $\x_{t}^k$ is processed using only the corresponding branch $G_{k}$, \ie $G_{k}(\x_{t}^k)$. Our multi-branch adversarial loss is an extension of the adversarial loss~\cite{Tsai_adaptseg_2018}, which aligns the prediction distributions of the source domain $G_{s}(\x_{s})$, and the sub-target domains $\{G_{k}(\x_{t}^{k})\}$. The multi-branch adversarial loss $\cL_{sadv}$ and the corresponding discriminator training loss $\cL_{sd}$ are formulated as,
\begin{eqnarray}
    \cL_{sadv}(G) =& -\sum_{k=1}^{K}\mathbb{E}_{\x_{t}^{k}\sim P_{T^{k}}}\log(D(G_{k}(\x_{t}^{k}))^{(n,1)}),
    \label{eq:multibranch_loss} \\
    \cL_{sd}(D) =& \!\!\!\!\!\!\!\!\!\!\!\!\!\!\!\!\!\!\!\!\!\!\!\!-\mathbb{E}_{\x_{s}\sim P_{S}}\log(D(G_{s}(\x_{s}))^{(n,1)}) \label{eq:loss_sd}\\
    &- \sum_{k=1}^{K}\mathbb{E}_{\x_{t}^{k}\sim P_{T^{k}}}\log(D(G_{k}(\x_{t}^{k}))^{(n,0)}), \nonumber
\end{eqnarray}
where $P_{S}$ and $P_{T^{k}}$ are the underlying data distributions of $\cS$ and $\cT_{k}$, respectively. The following full optimization objective is used for training our split module,
\begin{eqnarray}
    \cL_{split}(G) = \cL_{seg}(G) + \lambda_1 \cL_{sadv}(G),
    \label{eq:split_objective}
\end{eqnarray}
where $\lambda_1$ is a trades-off parameter. During the training process, we alternatively optimize the discriminator $D$ and the generator $G$ with the objective in the Eq. (\ref{eq:loss_sd}) and the Eq. (\ref{eq:split_objective}), respectively.

\subsection{Fuse: HyperNetwork for Branches Fusion} \label{sec:fuse_sec}
The \textbf{cluster} and \textbf{split} module discretizes the target domain into a few clusters, providing an initial discrete modeling of the target domain. The fuse of the discretized modes form continuous manifold, the sample on which reflects the continuous change of the target domain and might correspond to an unseen domain. In the \textbf{fuse} module, we learn to combine the sub-target domain to model the compound target domain continuously. 

\textbf{Compound Domain Modelling.} 
Here we model the target domain $\cT$ in the corresponding feature domain $\cF$, which is mapped by $F: \cT\rightarrow \cF$. Let $P(\widetilde{\f_{t}}^{k}|\x_{t}, k)$ 
be the feature distribution corresponding to image $\x_{t}$ when assumed to be from the $k^{\text{th}}$ cluster. 
Then the distribution of the feature $\widetilde{\f_{t}}$ of the image $\x_{t}$, \ie
$P(\widetilde{\f_{t}}|\x_{t})$, is expressed as, 
\begin{eqnarray}
    P(\widetilde{\f_{t}}|\x_{t}) \!=\!\!\! \sum_{k=1}^{K}P(\widetilde{\f_{t}}^{k}, k|\x_{t}) 
    \!=\!\frac{1}{N}\sum_{k=1}^{K}P(k|\x_{t})P(\widetilde{\f_{t}}^{k}|\x_{t}, k)\label{eq:commodel}
\end{eqnarray}
where $N = \int_{\widetilde{\f_{t}}^{k}}\sum_{k=1}^{K}P(\widetilde{\f_{t}}^{k}|\x_{t}, k)P(k|\x_{t})\text{d}\widetilde{\f_{t}}^{k}$. $P(k|\x_{t})$ describes the probability distribution of the sub-target domain's label of image  $\x_{t}$.
By taking the sub-target domain distributions $P(\widetilde{\f_{t}}^{k}|\x_{t}, k)$ as basis, the compound target domain can be modeled with the vector, \ie $\{[P(1|\x_{t}), ..., P(k|\x_{t}), ..., P(K|\x_{t})]'\}$.

\textbf{HyperNetwork for Branches Fusion.} In essence, the cluster and split module can be seen as modeling the sub-target domain label distribution as $P(k|\x_{t}) = 1, \text{if } E_{l}(\x_{t}) = k$ and $P(k|\x_{t}) = 0, \text{if } E_{l}(\x_{t}) \neq k$. 
It models the compound target domain as the discretized points in the vector space, as illustrated in Fig. \ref{fig:framestruct}\textcolor{red}{b}. 
In order to model the compound target domain in the continuous space, in our fuse module, we adopt the categorical distribution for $P(k|\x_{t})$, \ie
\begin{eqnarray}
    P(k|\x_{t}) = w_{k}, \,\,\,\,\,\, \text{ with, }\,\,\,\,\,\,
    \sum_{k=1}^{K} w_{k} = 1, w_{k}>0, 
\end{eqnarray}
where $\w = [w_{1}, ..., w_{k}, ..., w_{K}]^\top$ is the K-dimensional categorical vector, whose element $w_{k}$ represents the probability that the target image $\x_{t}$ belongs to the sub-target domain $\cT_k$. Then the hypernetwork $H(\cdot)$ is adopted to learn the $P(k|\x_{t})$, by taking the style code $\c_{t}$ of the image sample $\x_{t}$ as input, \ie
    $[w_{1}, ..., w_{k}, ..., w_{K}]^\top = H(\c_{t}).$
Substituting the $H(\c_{t})$ in Eq.~\eqref{eq:commodel}, the feature distribution $P(\widetilde{\f_{t}}|\x_{t})$ can be derived as,
\begin{eqnarray}
  P(\widetilde{\f_{t}}|\x_{t}) \sim \sum_{k=1}^{K} H(\c_{t})^{(k)}P(\widetilde{\f_{t}}^{k}|\x_{t}, k).
  \label{eq:fused_label}
\end{eqnarray}
where $H(\c_{t})^{(k)}$ is the $k^\text{th}$ element of $H(\c_{t})$. Eq.~(\ref{eq:fused_label}) shows that the compound target domain is modeled in the continuous vector space, $H(\c_{t})$, taking the sub-target domain distributions $P(\widetilde{\f_{t}}^{k}|\x_{t}, k)$ as basis, as illustrated in Fig. \ref{fig:framestruct}\textcolor{red}{b}. 

From above, it is shown that $H(\c_{t})$ weights the different sub-target domain distribution differently to get the compound target domain distribution. Here we adopt the network $G$ as our mapping $\cF$. Following~\cite{huang2007correcting}, we reweight each feature sample $\widetilde{\f_{t}}^{k} = G_k(\x_t)$ with $H(\c_{t})$, so that the feature sample from dominant sub-target domain has higher weight, whereas the sample from non-dominant sub-target domain has lower weight. The final prediction can be represented as,
\begin{eqnarray}
  \widetilde{\y_{t}} = \sum_{k=1}^{K} H(\c_{t})^{(k)}G_{k}(\x_{t}).
  \label{eq:sample_fuse}
\end{eqnarray}
By combining Eq. (\ref{eq:sample_fuse}) and Eq. (\ref{eq:multibranch_loss}), the adversarial loss for the fuse module $\cL_{fadv}$ and the corresponding discriminator training loss $\cL_{fd}$ can be formulated as,
\begin{eqnarray}
    \cL_{fadv}(G, H) = -\mathbb{E}_{\x_{t}\sim P_{T}}\log(D(\widetilde{\y_{t}})^{(n,1)}) \label{eq:fuse_adv_loss}\\
    \cL_{fd}(D) = -\mathbb{E}_{\x_{s}\sim P_{S}}\log(D(G_{s}(\x_{s}))^{(n,1)}) \label{eq:loss_fd}\\
    -\mathbb{E}_{\x_{t}\sim P_{T}}\log(D(\widetilde{\y_{t}})^{(n,0)}). \nonumber
    \label{eq:fuse_loss}
\end{eqnarray}
The optimization objective of our fuse module is a combination of Eq. (\ref{eq:seg_loss}) and Eq. (\ref{eq:fuse_adv_loss}),  which is given by, 
\begin{eqnarray}
    \cL_{fuse}(G, H) = \cL_{seg}(G) + \lambda_2 \cL_{fadv}(G, H),
    \label{eq:fuse_objective}
\end{eqnarray}
where $\lambda_2$ is the hyperparameter to blance between the adversarial loss and the segmentation loss. During the training process, we alternatively optimize the discriminator $D$ and the generator $G$, the hypernetwork $H$ with the objective in the Eq. (\ref{eq:loss_fd}) and the Eq. (\ref{eq:fuse_objective}), respectively. 
In our MOCDA model, the training of the fuse module is combined with the MAML strategy, which is explained further in Section \ref{sec:maml_update} and Algorithm \ref{alg:maml_train}.

\subsection{Update: MAML based Online Update} \label{sec:maml_update}
In the previous OCDA work \cite{liu2020open}, the open set is only treated as a testing set to verify the generalization ability of the model. In contrast, in our work, the open set is also used for updating the model online during testing, for better generalization to the unseen domain, realized by MAML. 

\textbf{MAML.} 
The MAML strategy \cite{finn2017model} aims at learning the optimal model parameters $\theta^*$, which eases the adaptation process for new tasks. In each iteration of MAML, there are two training loops; inner and outer. Let the data of inner and outer loops be $\cD_{in}$ and $\cD_{out}$, respectively. In each training iteration, the model parameters $\theta$ are first updated with the inner loop loss $\cL_{in}$ and data $\cD_{in}$. The updated model is then evaluated on the outer loop loss $\cL_{out}$ and data $\cD_{out}$, to test the generalization ability of the updated model. Furthermore, the evaluation performance $\cL_{out}$ is also adopted during update, to better generalize the model. This nested training fashion mimics the training and testing phase of the model. In order to endow adaptation ability, the optimization objective of MAML is formulated as,
\begin{eqnarray}
    \theta^* = \argmin_{\theta}\; \cL_{out}(\theta - \alpha \nabla \cL_{in}(\theta, \; \cD_{in}), \; \cD_{out}),
\end{eqnarray}
where $\alpha$ is the learning rate for updating the model.

\textbf{MAML for OCDA.} In our addressed problem of OCDA for semantic segmentation, images from the set $\{\x_{o}\}$ of the unseen open domain $\cO$  are available only during testing. We adopt the MAML algorithm in our MOCDA during training to be combined with the fuse module. MAML then offers us the advantage of quick adaptation to the open set during testing, by means of online update within one gradient step. 

In the inner loop,  we sample data from the target domain $\cT$, \ie $\cD_{in} = \{\x_{t}\}$. Meanwhile, in order to update the model without supervision, we use the unsupervised self-entropy loss\cite{vu2019advent} $\cL_{ent}$ as the inner loop loss $\cL_{in}$ -- which mimics the model update process during testing, given by, 
\begin{eqnarray}
\label{eq:ent_loss}
    \cL_{in} = \cL_{ent} =& \displaystyle-\frac{1}{HW}\sum_{n=1}^{HW}\sum_{c=1}^C \widetilde{\y_{t}}^{(n, c)} \log \widetilde{\y_{t}}^{(n, c)}.
\end{eqnarray}
In the outer loop, the data is sampled from both source domain $\cS$ and the target domain $\cT$, \ie $\cD_{out}= \{\x_{s}, \y_{s}, \x_{t}\}$. In order to evaluate the model's performance on different domains and in different way, the outer loop loss $\cL_{out}$ uses the optimization objective of the fuse module in Eq. (\ref{eq:fuse_objective}) and the self-entropy loss in Eq. (\ref{eq:ent_loss}), such that,
\begin{eqnarray}
\cL_{out} = \cL_{fuse} + \delta \cL_{ent},
\label{eq:out_loss}
\end{eqnarray}
where $\delta$ is the hyperparameter to balance between the fuse module loss and the unsupervised self-entropy loss.
The MAML algorithm used during OCDA training is presented in Algorithm \ref{alg:maml_train}. Similarly, the MAML used during the online update, of OCDA  testing, is given in Algorithm~\ref{alg:maml_test}.

\subsection{Training Protocol of MOCDA}
In total, our MOCDA model is trained in the multi-stage way, consisting of three steps: i) training the MUNIT model for style code extraction and clustering, ii) training with the CDBN layer in split module, iii) the CDBN layer is frozen, adding the hyper-network and the fuse module, and training the hypernetwork $H$ and fine-tuning the semantic segmentation network $G$ with MAML strategy as described in Algorithm \ref{alg:maml_train}. Then during testing stage, our whole model, except for CDBN layer, is online updated with the MAML strategy as clarified in Algorithm~\ref{alg:maml_test}.

\begin{algorithm}[!t]
\caption{MAML algorithm for OCDA (Training)}
\textbf{Require:} Source data $\cS=\{(\x_{s}, \y_{s}) \}$,  target data ${\cT=\{\x_{t} \}}$, segmentation network $G$, hypernetwork $H$, discriminator $D$, the learning rate $\alpha$ of $G, H$, and the learning rate $\zeta$ of discriminator $D$.
\begin{algorithmic}[1]
\State Initialize the parameters $\theta_{GH}$ and $\theta_{D}$, respectively of the segmentation network $G$, hypernetwork $H$, and the discriminator $D$;
\While{not done}
\State Sample $\cD_{in}$ from $\cT$ \algorithmiccomment{Inner Loop}
\State $\theta_{GH}^{+}\leftarrow \theta_{GH}-\alpha \nabla_{\theta_{GH}} \cL_{in}(\cD_{in}, \theta_{GH})$;
\State Sample $\cD_{out}$ from $\cS$ and $\cT$ \algorithmiccomment{Outer Loop}
\State $\theta_{GH}\leftarrow\theta_{GH}-\alpha \nabla_{\theta_{GH}} \cL_{out}(\cD_{out}, \theta_{GH}^{+})$;
\State $\theta_{D}\leftarrow\theta_{D}-\zeta \nabla_{\theta_{D}} \cL_{fd}(\cD_{out}, \theta_{D})$;
\EndWhile
\end{algorithmic}
\label{alg:maml_train}
\end{algorithm}

\begin{algorithm}[!t]
\caption{MAML algorithm for OCDA (Testing)}
\textbf{Require:} Data $\{\x_{o}\}$ from the unseen novel domain $\cO$,  segmentation network $G$, hypernetwork $H$.
\begin{algorithmic}[1]
\State Use trained parameters $\theta_{GH}$ of the segmentation network, $G$ and the hypernetwork $H$, from the training phase;
\State $F\leftarrow 0$
\For{$i=1, ..., n$}
\State Sample the $i^\text{th}$ image  $\x_{o}^{i}$ from $\{\x_{o}\}$;
\State $\widetilde{\y_{o}^{i}}\leftarrow G(\x_{o}^{i})$;
\State $\theta_{GH}\leftarrow \theta_{GH}-\eta\nabla_{\theta_{GH}}\cL_{ent}(\widetilde{\y_{o}^{i}}, \theta_{GH})$
\EndFor
\end{algorithmic}
\label{alg:maml_test}
\end{algorithm}

%% file: 05_Experimetns.tex
\section{Experiments}
In this section, we demonstrate the benefits of our MOCDA model under the open compound domain adaptive semantic segmentation setting. We compare our MOCDA model with other state-of-the-art (SOTA) methods on both of the target domain and the open domain. In order further prove the effectiveness of our MOCDA model for open domain with online update, we introduce more diverse and challenging extended open domains to test the model performance additionally.

\subsection{Experiments Setup}
Following \cite{liu2020open}, we adopt the synthetic image dataset GTA5 \cite{richter2016playing} or SYNTHIA-SF \cite{sankaranarayanan2017unsupervised} as the source domain, the rainy, snowy, and cloudy images in BDD100K\cite{yu2020bdd100k} as the target domain, while the overcast images in BDD100K are utilized as the open domain. 
Besides, more diverse images from other real image datasets, Cityscapes\cite{cordts2016cityscapes}, KITTI\cite{Alhaija2018IJCV} and WildDash \cite{Zendel_2018_ECCV} are introduced as extended open domains. We adopt the the DeepLab-VGG16 model \cite{chen2017deeplab, simonyan2014very} with the batch normalization layer as the segmentation network. The cluster numbers $K$ is set as 4. The semantic segmentation network and discriminator structure is the same as \cite{Tsai_adaptseg_2018}. The hyperparameter $\lambda_1$ and $\lambda_2$ in Eq.(\ref{eq:fuse_objective}) and Eq.(\ref{eq:split_objective}) are set as 0.001. The hyperparameter $\delta$ in Eq.(\ref{eq:out_loss}) is set as 0.0001. More detailed introduction of the dataset and the implementation details of our model are put in the supplementary due to the space limit.
\begin{figure}
    \centering
    \includegraphics[width=\linewidth]{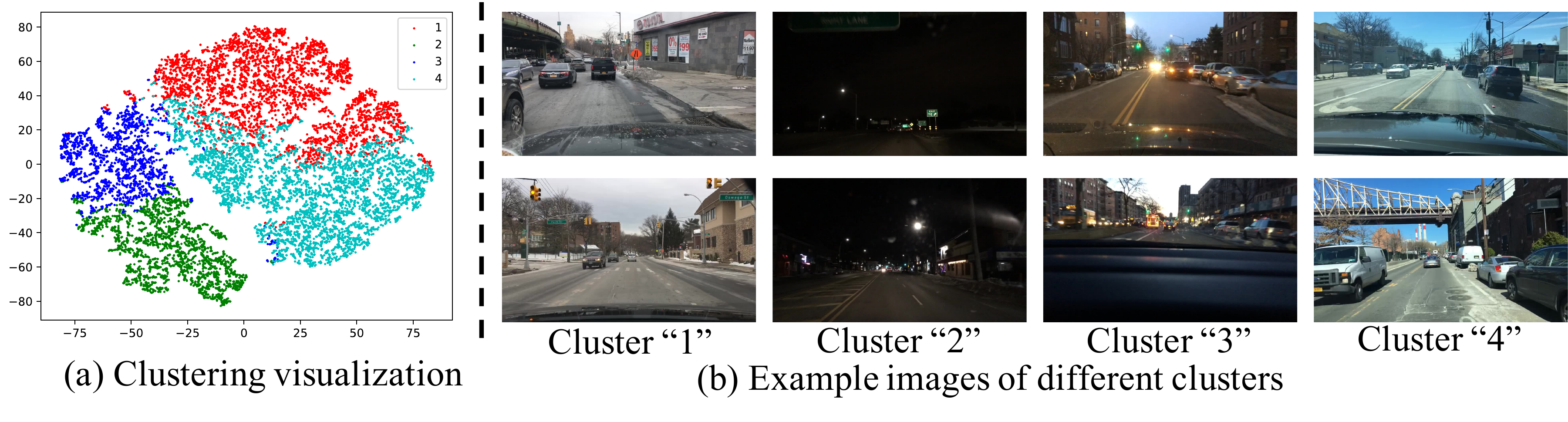}
    \caption{Visualization of clustering results. (a) is the t-SNE visualization of the style code extracted by the cluster module, (b) is example images from different clusters.}
    \label{fig:cluster_vis}
\end{figure}
    
\subsection{GTA5 to BDD100K}
\begin{figure}
    \centering
    \includegraphics[width=\linewidth]{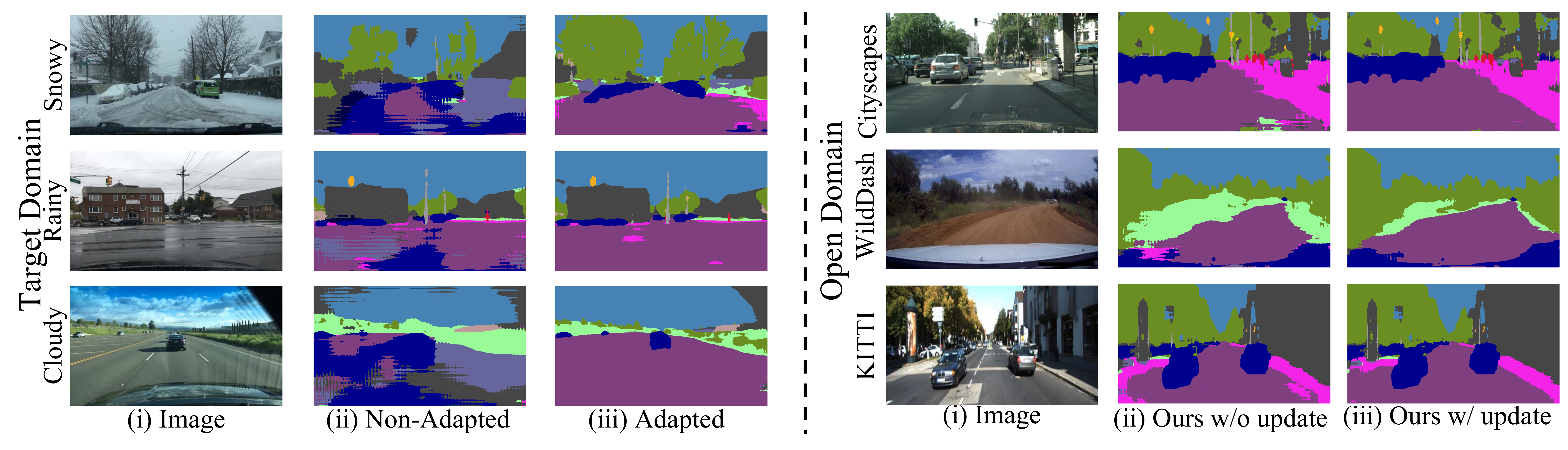}
    \caption{Qualitative comparison of semantic segmentation results on the target domain, including the rainy, snowy and cloudy weather, and on the open domains, KITTI, WildDash and Cityscapes.}
    \label{fig:target_seg_vis}
\end{figure}

\textbf{Comparison with SOTA.} In Table \ref{tab:comp_sota_gta}, we present our open compound domain adaptation results, in comparison with other SOTA methods. For fair comparison, all of the methods adopt the DeepLab-VGG16 model with the batch normalization layer. Compared with our baseline method AdaptSegNet\cite{Tsai_adaptseg_2018}, our split module achieves $3.1\%$ and $2.4\%$ gain on the target domain and the open domain, respectively. Compared with the SOTA method OCDA\cite{liu2020open}, our split module performance outperforms by $0.9\%$ on the target domain and by $1.6\%$ on the open domain. It proves the effectiveness of our cluster module and the split module, for sub-target domain discovery and sub-target domain-specific information disjointing. The clustering visualization is shown in Fig. \ref{fig:cluster_vis}. Then by adopting the meta-learning with the hypernetwork and the MAML training strategy in the fuse module, our MOCDA model achieves the state-of-the-art performance, which improves the split module performance by $2.3\%$ from $25.4\%$ to $27.7\%$, and by $1.9\%$ from $29.5\%$ to $31.4\%$ on the target domain and the open domain, respectively. It proves the advantage of our MOCDA model on fusing the different sub-target domains knowledge, modeling the target domain continuously through the hypernetwork, and adopting the MAML training strategy. The qualitative comparison of the semantic segmentation results on the target domain is shown in Fig. \ref{fig:target_seg_vis}.
\begin{table}
\centering
\resizebox{0.48\textwidth}{19.5mm}{
\begin{tabular}{c|ccc|c|cc}
 \hline
 
 \hline
  {Source}&\multicolumn{3}{c|}{Compound}&Open&\multicolumn{2}{c}{Avg}\\
  GTA$\rightarrow$ & Rainy & Snowy & Cloudy & Overcast & C & C+O\\
  \hline
  \hline
  Source Only\cite{liu2020open} & 16.2 & 18.0 & 20.9 & 21.2 & 18.9 & 19.1\\
  Source Only $^\ast$ & 19.7 & 18.4 & 20.5 & 22.5 & 19.7 & 21.0\\
  \hline
  AdaptSegNet\cite{liu2020open} & 20.2 & 21.2 & 23.8 & 25.1 & 22.1 & 22.5\\
  AdaptSegNet\cite{Tsai_adaptseg_2018} $^\ast$ & 21.6 & 20.5 & 23.9 & 27.1 & 22.3 & 24.4\\
  CBST\cite{zou2018unsupervised} & 21.3 & 20.6 & 23.9 & 24.7 & 22.2 & 22.6 \\
  IBN-Net\cite{pan2018two} & 20.6 & 21.9 & 26.1 & 25.5 & 22.8 & 23.5\\
  PyCDA \cite{lian2019constructing} & 21.7 & 22.3 & 25.9 & 25.4 & 23.3 & 23.8 \\
  OCDA \cite{liu2020open} & 22.0 & 22.9 & 27.0 & 27.9 & 24.5 & 25.0 \\
  \hline
  Ours (Split) &23.5 &23.5 &27.8 & 29.5 & 25.4 &27.1 \\
  Ours (Fuse) &\textbf{24.4}&\textbf{27.5}& \textbf{30.1}& \textbf{31.4}& \textbf{27.7}& \textbf{29.4}\\
  \hline
  
  \hline
 \end{tabular}
 }
\captionof{table}{\label{tab:comp_sota_gta}Semantic segmentation performance comparison with SOTA: GTA$\rightarrow$ BDD100K with DeepLab-VGG16 backbone. The results are reported on mIoU over 19 classes. $^\ast$ means our reproduced result. The best results are denoted in bold.}
\end{table}

\textbf{Online Update.} Another meta-learning paradigm in our MOCDA model, besides the fuse module, is the MAML algorithm based online update during testing stage. 
From Table \ref{tab:online_up_gta}, it is shown that our MOCDA model without online update outperforms the baseline method AdaptSegNet \cite{Tsai_adaptseg_2018} on both of the open domain and the extended open domain by $5.6\%$ in average. It proves the effectiveness of our cluster, split and fuse module for open domain generalization. By further using the MAML based online update strategy described in Algorithm \ref{alg:maml_test} during the testing stage, our MOCDA model performance on all the open domains improves by $0.7\%$ in average, from $28.1\%$ to $28.8\%$. Our model w/ or w/o online update has the same performance on the open domain, BDD100K overcast image. It is due to that the BDD100K overcast image is still from the BDD100K dataset, and the style gap between the overcast image and the target domain image is very narrow, whose visualization is shown in supplementary. The benefit from our cluster, split and fuse module has been already able to handle the narrow style gap and have good generalization performance already. The performance gain, $0.7\%$, $1.1\%$ and $1.0\%$ on the extended open domains where the style gap is much larger, Cityscapes, KITTI and WildDash dataset, proves that the MAML based meta-learning paradigm, in Algorithm \ref{alg:maml_train} for training and Algorithm \ref{alg:maml_test} for testing, endows the fast adaptation ability to our model to generalize better on the open domains. The qualitative comparison, w/ or w/o online update, of the semantic segmentation results on the open domains are shown in Fig. \ref{fig:target_seg_vis}.
\begin{table}
\centering
\resizebox{0.48\textwidth}{15mm}{
\begin{tabular}{c|c|ccc|c}
 \hline
 
 \hline
  {Source}&Open&\multicolumn{3}{c|}{Extended Open}&\multirow{2}{*}{Avg}\\
  GTA$\rightarrow$ & BDD & Cityscapes & KITTI & WildDash\\
  \hline
  \hline
  Source\cite{liu2020open} & 21.2 & -- & -- & -- & --\\
  Source$^\ast$ & 22.5 & 19.3 & 24.1 & 16.0 & 20.5\\
  AdaptSegNet\cite{liu2020open} & 25.1 & -- & -- & -- & -- \\
  AdaptSegNet\cite{Tsai_adaptseg_2018} $^\ast$ & 27.1 & 22.0 & 23.4 & 17.5 & 22.5\\
  w/o Online Update & 31.4 & 30.4 & 29.8 & 20.6 & 28.1\\
  w/ Online Update & 31.4 & 31.1 & 30.9 & 21.6 & \textbf{28.8}\\
  Gain of Online Update & -- & \textcolor{red}{+0.7} & \textcolor{red}{+1.1} & \textcolor{red}{+1.0} & \textcolor{red}{+0.7}\\
  \hline
  
  \hline
 \end{tabular}
 }
\captionof{table}{\label{tab:online_up_gta}Open domain semantic segmentation performance comparison w/ or w/o online update: GTA$\rightarrow$ BDD100K with DeepLab-VGG16 backbone. The results are reported on mIoU over 19 classes. $^\ast$ means our reproduced result. The best results are denoted in bold.}
\end{table}

\textbf{Ablation Study.} In order to verify the effectiveness of different components of our MOCDA model, we show the performance comparison of ablations and different variants of our model in Table \ref{tab:ablation}. From Table \ref{tab:ablation}, it is shown that all the modules, the cluster/split module ($\cL_{split}$), the fuse module ($\cL_{fadv}$) and the MAML training strategy are helpful to our whole MOCDA model. The cluster and split module has been proven to be helpful in the comparison with AdaptSegNet\cite{Tsai_adaptseg_2018} and other SOTA methods. Here we show the effectiveness of our meta-learning paradigm, the hypernetwork and the MAML training strategy through the ablations and variants methods comparison. 
Firstly, in order to prove the validity of our hypernetwork, we build the baseline methods of the branch fusion in non-adaptive way; 1), averagely fuse for prediction during the testing stage of the split module. 2), averagely fuse during the training and testing stage of the fuse module. 3) use the style code distance from different clusters to weight different branches during the training and testing stage of the fuse module. It is shown that our hypernetwork based branch fusion strategy performance, $27.1\%$, outperforms all other non-adaptive fusion strategy, $23.1\%, 26.1\%, 26.6\%$. It benefits from the advantage of adaptive weights predicted from the hypernetwork conditioned on the image sample style code. And the t-SNE visualization of hypernetwork prediction is shown in supplementary to prove the validity of the hypernetwork prediction. Secondly, by comparing the performance of training the fuse module using the $\cL_{out}$ in the Eq. (\ref{eq:out_loss}) and purely using the $\cL_{fuse}$ in Eq.(\ref{eq:fuse_objective}), it is shown that there is $0.2\%$ performance gain by adding the unsupervised entropy loss, from $27.1\%$ to $27.3\%$. By further introduce the MAML training strategy in Algorithm \ref{alg:maml_train} for the fuse module, as done in our MOCDA model, the performance can be further improved to $27.7\%$. It proves that the MAML training strategy is not only helpful to the open domain generalization as described above, but also is beneficial to improve the adaptation performance of the model on the target domain. It results from that MAML training strategy mimics the training and testing procedure with the outer loop and inner loop and makes the model more domain adaptive.
\begin{table}
\centering
\resizebox{0.4\textwidth}{18mm}{
\begin{tabular}{cccccc|c}
 \hline
  $\cL_{seg}$&$\cL_{adv}$&$\cL_{sadv}$&$\cL_{fadv}$&$\cL_{ent}$&MAML&mIoU\\
  \hline
  \checkmark&&&&&&18.9\\
  \checkmark&\checkmark&&&&&22.3\\
  \checkmark&&\checkmark&&&&25.4\\
  \checkmark&&\checkmark&&&&23.1$^{\dagger}$\\
  \checkmark&&&\checkmark&&&26.1$^{\ddagger}$\\
  \checkmark&&&\checkmark&&&26.6$^{\mathsection}$\\
  \checkmark&&&\checkmark&&&27.1\\
  \checkmark&&&\checkmark&\checkmark&&27.3\\
  \checkmark&&&\checkmark&\checkmark&\checkmark&\textbf{27.7}\\
  \hline
 \end{tabular}
 }
\caption{\label{tab:ablation} Different ablations and variants comparison for OCDA, tested on BDD100k target domain based on DeepLab-VGG16 with batch normalization layer backbone. The results are reported on mIoU over 19 classes. $^{\dagger}$ represents the average fusion only during testing. $^{\ddagger}$ represents the average fusion of different branches during training and testing. $^{\mathsection}$ represents the style code distance weighted fusion during training and testing.}
\end{table}

\subsection{SYNTHIA-SF to BDD100K}
In this section, SYNTHIA-SF is used as the source domain. Following \cite{zhou2020domain}, we only take 11 main classes in the SYNTHIA-SF dataset to measure the semantic segmentation performance, which are road, sidewalk, building, wall, fence, pole, light, vegetation, sky, person and car.

\textbf{Comparison with SOTA.} In Table \ref{tab:comp_sota_synsf}, we report the quantitative comparison results between our MOCDA model and other SOTA methods for the open compound domain adaptation setting, from the SYNTHIA-SF to the BDD100K. From Table \ref{tab:comp_sota_synsf}, it is shown that our MOCDA model outperforms the MinEnt \cite{vu2019advent} method by $6.5\%$ and by $6.9\%$ on the target and open domain, respectively. Meanwhile, compared with the AdaptSegNet method \cite{Tsai_adaptseg_2018}, our MOCDA model has performance gain of $2.4\%$ and $2.3\%$ on the target domain and the open domain, respectively. It further verifies the effectiveness of our MOCDA model for open compound domain adaptation.

\begin{table}
\centering
\resizebox{0.48\textwidth}{12.5mm}{
\begin{tabular}{c|ccc|c|cc}
 \hline
 
 \hline
  {Source}&\multicolumn{3}{c|}{Compound}&Open&\multicolumn{2}{c}{Avg}\\
  SYNTHIA-SF$\rightarrow$ & Rainy & Snowy & Cloudy & Overcast & C & C+O\\
  \hline
  \hline
  Source Only & 16.5 & 18.2 & 21.4 & 20.6 & 19.2 & 19.8\\
  \hline
  MinEnt\cite{vu2019advent} & 21.8 & 22.6 & 26.2 & 25.7 & 23.9 & 24.7\\
  AdaptSegNet\cite{Tsai_adaptseg_2018} & 24.9 & 26.9 & 30.7 & 30.3 & 28.0 & 29.0\\
  \hline
  Ours (Split) & 25.2 & 27.9 & 32.4 & 31.8 &  29.1& 30.3\\
  Ours (Fuse) & \textbf{26.6} & \textbf{30.0} & \textbf{33.0} & \textbf{32.6} & \textbf{30.4}& \textbf{31.4}\\
  \hline
  
  \hline
 \end{tabular}
 }
\captionof{table}{\label{tab:comp_sota_synsf}Semantic segmentation performance comparison with SOTA: SYNTHIA-SF$\rightarrow$ BDD100K with DeepLab-VGG16 backbone. The results are reported on mIoU over 11 classes. The best results are denoted in bold.}
\end{table}

\textbf{Online Update.} In Table \ref{tab:comp_up_synsf}, the performance of our MOCDA model for the open domain and the extended open domain are shown. Our MOCDA model w/o online update outperforms the AdaptSegNet method by $2.2\%$ in average on all the open domains. By further utilizing the online update in the open domain, the peformance can be further improved by $1.1\%$ in average, from $30.1\%$ to $31.2\%$. It further proves the validity of the online update for the open domain.
\begin{table}
\centering
\resizebox{0.48\textwidth}{11.7mm}{
\begin{tabular}{c|c|ccc|c}
 \hline
 
 \hline
  {Source}&Open&\multicolumn{3}{c|}{Extended Open}&\multirow{2}{*}{Avg}\\
  SYNTHIA-SF$\rightarrow$ & BDD & Cityscapes & KITTI & WildDash\\
  \hline
  \hline
  Source & 20.6 & 24.7 & 20.7 & 17.3 & 20.8\\
  AdaptSegNet\cite{Tsai_adaptseg_2018} & 30.3 & 35.9 & 24.7 & 20.7 & 27.9\\
  w/o Online Update & 32.6 & 29.9 & 33.2 & 24.5 & 30.1\\
  w/ Online Update & 32.6 & 32.2 & 34.2 & 25.8 & \textbf{31.2}\\
  Gain of Online Update & -- & \textcolor{red}{+2.3} & \textcolor{red}{+1.0} & \textcolor{red}{+1.3} & \textcolor{red}{+1.1}\\
  \hline
  
  \hline
 \end{tabular}
 }
\captionof{table}{\label{tab:comp_up_synsf}Open domain semantic segmentation performance comparison w/ or w/o online update: SYNTHIA-SF$\rightarrow$ BDD100K with DeepLab-VGG16 backbone. The results are reported on mIoU over 11 classes. $^\ast$ means our reproduced result. The best results are denoted in bold.}
\end{table}

%% file: 06_Conclusion.tex
\section{Conclusion}
In this paper, we address the problem of open compound domain adaptation, and propose a meta-learning based model, MOCDA. MOCDA is composed of four modules, cluster, split, fuse and update module. Meta-learning serves in the fuse and update module for continuously modeling the compound target domain and online update. The extensive experiments show that our model achieves the state-of-the-art performance on different benchmarks, proving the effectiveness of our proposed MOCDA model.

%% file: 08_Supplementary.tex
\renewcommand{\thesection}{S\arabic{section}}  
\renewcommand{\thefigure}{S\arabic{figure}}
\renewcommand{\thetable}{S\arabic{table}}
\setcounter{section}{0}
\setcounter{table}{0}
\setcounter{figure}{0}
\section*{Supplementary}
In this supplementary, we provide additional information for,
\begin{itemize}
    \item[\textbf{S1}] implementation details of our MOCDA model,
    \item[\textbf{S2}] more detailed information about the datasets in our experiments,
    \item[\textbf{S3}] additional experimental results and qualitative results on the OCDA benchmark,
    \item[\textbf{S4}] additional visualization results for the style code and hypernetwork prediction.
\end{itemize}

\section{Detailed Implementation of our MOCDA model}
In the main paper, we introduce our MOCDA model in the Sec. \textcolor{red}{3} and the implementation details in the Sec. \textcolor{red}{4.1}. Here we provide more detailed implementation of different modules in our MOCDA model, separately. 

\textbf{Cluster.} In the cluster module, we train the MUNIT \cite{huang2018multimodal} model to translate between the source domain images and the compound target domain images in the unsupervised way. We follow the experimental set up in the urban scene image translation set up in MUNIT \cite{huang2018multimodal}. The shortest side of the images are firstly resized to 512, and then the images are randomly cropped with the size of $400\times 400$. The loss weights for image reconstruction loss, style reconstruction loss, content reconstruction loss, and domain-invariant perceptual loss are set as 10, 1, 1, and 1, respectively. The Adam optimizer \cite{kingma2014adam} is adopted with $\beta_1=0.5, \beta_2= 0.999$, and the learning rate is set as 0.0001. Also, the dimension of the style code is set as 8. The number of the clusters $K$ is set as 4. 

\textbf{Split, Fuse, and Update.} In the split and fuse module, we have the semantic segmentation network and the discriminator. We adopt the DeepLab-VGG16 \cite{chen2017deeplab, simonyan2014very} with synchronized batch normalization layer \cite{ioffe2015batch} for the semantic segmentation network. And we adopt the discriminator structure in \cite{Tsai_adaptseg_2018}. The compound target domain images and the open domain images, from BDD100K \cite{yu2020bdd100k}, Cityscapes\cite{cordts2016cityscapes}, WildDash \cite{Zendel_2018_ECCV} and KITTI \cite{Alhaija2018IJCV}, are resized to $1024\times 512$, and the source domain images from GTA5 \cite{richter2016playing} and SYNTHIA-SF \cite{HernandezBMVC17} are resized to $1280\times 720$. The $\lambda_1$ in Eq. (\textcolor{red}{7}), and $\lambda_2$ in Eq. (\textcolor{red}{14}) of the main paper are set as 0.001. In the update module, during the training stage, the $\delta$ in Eq. (\textcolor{red}{17}) is set as 0.0001. 
In the split, fuse and update module, we adopt the SGD optimizer to train the hypernetwork and the semantic segmentation network, where the momentum is 0.9 and the weight decay is $5\times 10^{-4}$. The learning rate is set as $2.5\times 10^{-4}$, and uses the polynomial decay strategy with power of 0.9 as done in \cite{Tsai_adaptseg_2018}. We keep the same learning rate for online updating the hypernetwork and the semantic segmentation network. Also, we adopt the Adam optimizer \cite{kingma2014adam} for training the discriminator with $\beta_1=0.9, \beta_2=0.99$. The learning rate is set as $1.0\times 10^{-4}$ and uses the polynomial decay strategy with power of 0.9. And our MOCDA model is implemented with PyTorch \cite{paszke2019pytorch}. 

\section{Datasets Overview}
In Sec. \textcolor{red}{4} of the main paper, we introduce the experiments setup of the OCDA benchmark, and there are six datasets in total, GTA5 \cite{richter2016playing}, SYNTHIA-SF \cite{HernandezBMVC17}, BDD100K \cite{yu2020bdd100k}, Cityscapes \cite{cordts2016cityscapes}, WildDash \cite{Zendel_2018_ECCV} and KITTI \cite{Alhaija2018IJCV}, involved in the experiments. Here we provide detailed information of involved datasets.

\textbf{GTA5.} GTA5 \cite{richter2016playing} is a synthetic urban scene image dataset, rendered from game engine. The scene of the GTA5 images is based on the city of Los Angeles. The GTA5 dataset covers 24966 densely labeled images, the annotation of which is compatible with that of Cityscapes. In OCDA benchmark, GTA5 $\rightarrow$ BDD100K, the GTA5 images, with the ground truth label, serve as source domain. 

\textbf{SYNTHIA-SF.} SYNTHIA-SF \cite{HernandezBMVC17} is a synthetically rendered image dataset from virtual city. There are 2224 images in the SYNTHIA-SF dataset, featuring different scenarios and traffic conditions. The images are densely labeled and the labels are compatible with Cityscapes. In our OCDA benchmark, SYNTHIA-SF $\rightarrow$ BDD100K, the SYNTHIA-SF dataset and the associated ground truth label serve as the source domain.

\textbf{BDD100K.} BDD100K \cite{yu2020bdd100k} is a real urban scene image dataset, mainly taken from US cities. And the images in BDD100K dataset are diverse in different aspects such as weather and environment. We adopt the C-driving subset of BDD100K proposed in \cite{liu2020open}, which is composed of rainy, snowy, cloudy and overcast images. During training stage, 14697 images, without the ground truth label, are used as the unlabeled compound target domain, including rainy, snowy and cloudy weather images. All different weather images are mixed and not assigned the weather information. During the testing stage, 803 images covering rainy, snowy and cloudy weather, with ground truth semantic annotation, are used as the validation set of the compound target domain, for evaluating the adaptation performance of the model. Besides, during the testing stage, 627 images with the ground truth semantic label, containing overcast weather, are taken as the validation set of the open domain, for evaluating the generalization performance of the model. The semantic label of the BDD100K dataset is compatible with that of Cityscapes.

\textbf{Cityscapes.} Cityscapes \cite{cordts2016cityscapes} is a real street scene image dataset, collected from different European cities. In our OCDA benchmark, during the testing stage, the validation set of Cityscapes, covering 500 densely labeled images, is used as one of the extended open domains to evaluate the generalization ability of the model.

\textbf{KITTI.} KITTI \cite{Alhaija2018IJCV} covers the real urban scene images, taken from the mid-size European city, Karlsruhe. In our OCDA benchmark, the validation set of KITTI, including 200 densely labeled images, is used as one of the extended open domains for generalization ability evaluation during the testing stage. The ground truth label of KITTI dataset is compatible with that of Cityscapes.

\textbf{WildDash.} WildDash \cite{Zendel_2018_ECCV} is a dataset covering images from diverse driving scenarios under the real-world conditions. The images in WildDash possess the diversity in different aspects, such as the time, weather, data sources and camera characteristics. In our OCDA benchmark, during the testing stage, the validation set of WildDash, containing 70 Cityscapes annotation compatible images, serves as one of the extended open domains for measuring the generalization performance of the model.

\section{Additional Experimental Results}
In Sec. \textcolor{red}{4} of the main paper, we provide the quantitative and qualitative experimental results of our MOCDA model on the OCDA benchmark. Here we provide the detailed quantitative experimental results, and additional qualitative experimental results. 

\textbf{Quantitative results.} In Table \textcolor{red}{1} and Table \textcolor{red}{4} of the main paper, the quantitative experimental results of our MOCDA model are reported on the mean IoU, for the OCDA task. Correspondingly, in Table \ref{tab:gtaperclass_opentarget} and Table \ref{tab:synsf_perclass_opentarget}, the more detailed per-class IoU results, on the compound target domain and the open domain, are shown. Additionally, the quantitative experimental results on different weather images are reported in Table \ref{tab:gtaperclass_target}. The detailed quantitative experimental results further verify the effectiveness of our MOCDA model for the OCDA task, on both of the compound target domain and the open domain.

\begin{table*}[!ht]
    \centering
    \resizebox{\textwidth}{!}{
    \begin{tabular}{c|c|cccccccccccccccccc|c}
    \hline
    
    \hline
        \multicolumn{21}{c}{GTA5$\rightarrow$BDD100K}\\
        \hline
        Domain & Method&\rotatebox{90}{road}&\rotatebox{90}{sidewalk}&\rotatebox{90}{building}&\rotatebox{90}{wall}&\rotatebox{90}{fence}&\rotatebox{90}{pole}&\rotatebox{90}{traffic light}&\rotatebox{90}{traffic sign}&\rotatebox{90}{vegetation}&\rotatebox{90}{terrian}&\rotatebox{90}{sky}&\rotatebox{90}{person}&\rotatebox{90}{rider}&\rotatebox{90}{car}&\rotatebox{90}{truck}&\rotatebox{90}{bus}&\rotatebox{90}{train}&\rotatebox{90}{motorbike}&mIoU\\
        \hline
        \multirow{4}{*}{Target} & 
        Source$^\ast$ & 32.1 & 12.4 & 47.1 & 3.9 & 22.6 & 17.6 & 9.9 & 4.7 & 52.0 & 13.9 & 74.6 & 24.3 & 0.0 & 38.0 & 10.0 & 10.4 & 0.0 & 0.0 & 19.7\\
        & AdaptSegNet$^\ast$\cite{Tsai_adaptseg_2018} & 46.9 & 14.0 & 60.2 & 5.9 & 20.4 & \textbf{18.3} & 9.0 & 4.6 & 48.9 & 14.1 & 78.2 & 24.6 & 0.0 & 48.7 & 13.1 & 16.5 & 0.0 & 0.0& 22.3\\
        & Ours(Split) & 71.6 & 13.4 & \textbf{63.7} & 8.2 & 19.9 & 18.2 & 6.8 & 5.6 & 57.3 & 16.5 & 80.9 & 22.7 & 0.0 & 57.4 & 18.7 & 21.2 & 0.0 & 0.0 & 25.4 \\
        & Ours (Fuse) & \textbf{73.9} & \textbf{20.6} & 58.2 & \textbf{8.5} & \textbf{22.8} & 17.9 & \textbf{10.4} & \textbf{7.1} & \textbf{61.9} & \textbf{20.1} & \textbf{84.8} & \textbf{26.1} & \textbf{2.3} & \textbf{61.3} & \textbf{19.8} & \textbf{26.4} & 0.0 & \textbf{3.7} & \textbf{27.7}\\
        \hline
        \multirow{4}{*}{Open$^\dagger$} & 
        Source$^\ast$ & 28.7 & 20.3 & 50.3 & 6.3 & 25.1 & 20.6 & 8.7 & 12.3 & 62.0 & 20.3 & 79.4 & 33.4 & 4.6 & 38.8 & 10.4 & 7.0 & 0.0 & 0.3 & 22.5\\
        & AdaptSegNet$^\ast$\cite{Tsai_adaptseg_2018} & 58.7 & 22.9 & 64.1 & 10.4 & 24.0 & \textbf{21.8} & 8.1 & 10.8 & 62.8 & 22.4 & 84.9 & \textbf{35.5} & 8.8 & 53.2 & 15.5 & 10.1 & 0.5 & 0.5 & 27.1\\
        & Ours (Split) & 76.5 & 22.0 & \textbf{68.6} & \textbf{15.8} & 22.6 & 21.6 & 6.0 & 6.8 & 64.8 & 24.3 & 86.6 & 35.2 & 8.1 & 63.1 & \textbf{26.1} & 11.7 & 0.1 & 0.0 & 29.5\\
        & Ours (Fuse) & \textbf{80.1} & \textbf{28.6} & 66.0 & 13.0 & \textbf{26.6} & 20.9 & \textbf{8.9} & \textbf{15.5} & \textbf{67.0} & \textbf{25.1} & \textbf{87.7} & 33.2 & \textbf{9.5} & \textbf{69.2} & 23.0 & \textbf{18.3} & \textbf{2.2} & \textbf{2.0} & \textbf{31.4}\\
   \hline
   
   \hline
    \end{tabular}
    }
    \caption{Per-Class IoU on the compound target domain and open domain of the OCDA benchmark: GTA5 $\rightarrow$ BDD100K. $\ast$ represents our reproduced result of the experiments in \cite{liu2020open}. The results are reported over 19 classes. The 'bicycle' class is not listed due to the result is close to zero. The best results are denoted in bold. Open $^\dagger$ is open domain covering the BDD100K overcast images.}
    \label{tab:gtaperclass_opentarget}
\end{table*}

\begin{table*}[!ht]
    \centering
    \resizebox{0.8\textwidth}{!}{
    \begin{tabular}{c|c|ccccccccccc|c}
    \hline
    
    \hline
        \multicolumn{14}{c}{SYNTHIA-SF$\rightarrow$BDD100K}\\
        \hline
        Domain & Method&\rotatebox{90}{road}&\rotatebox{90}{sidewalk}&\rotatebox{90}{building}&\rotatebox{90}{wall}&\rotatebox{90}{fence}&\rotatebox{90}{pole}&\rotatebox{90}{traffic light}&\rotatebox{90}{vegetation}&\rotatebox{90}{sky}&\rotatebox{90}{person}&\rotatebox{90}{car}&mIoU\\
        \hline
        \multirow{5}{*}{Target} & 
        Source & 3.1 & 6.8 & 42.7 & 0.0 & 0.0 & 10.2 & 1.1 & \textbf{39.6} & 69.2 & 9.7 & 28.2 & 19.2\\
        & MinEnt\cite{vu2019advent} & \textbf{67.2} & 1.8 & 50.7 & 0.0 & 0.0 & 4.4 & 1.3 & 11.7 & 71.8 & 8.7 & 45.7 & 23.9 \\
        & AdaptSegNet\cite{Tsai_adaptseg_2018} & 63.1 & 11.9 & 46.5 & 0.1 & 0.0 & 10.5 & 3.1 & 22.2 & \textbf{78.7} & 17.8 & 54.1 & 28.0\\
        & Ours(Split) & 59.8 & 15.5 & \textbf{52.8} & 0.2 & 0.0 & \textbf{13.6} & 2.3 & 28.4 & 73.3 & 19.2 & 55.1 & 29.1\\
        & Ours (Fuse) & 61.3 & \textbf{17.3} & 49.7 & \textbf{1.0} & \textbf{0.1} & 11.1 & \textbf{5.9} & 37.5 & 72.6 & \textbf{21.5} & \textbf{56.3} & \textbf{30.4}\\
        \hline
        \multirow{5}{*}{Open$^\dagger$} & 
        Source & 1.9 & 9.0 & 43.4 & 0.0 & 0.0 & 11.1 & 1.2 & \textbf{45.1} & 74.7 & 13.0 & 27.2 & 20.6 \\
        & MinEnt\cite{vu2019advent} & 68.9 & 2.5 & 51.6 & 0.0 & 0.0 & 5.7 & 1.4 & 14.2 & 77.2 & 11.7 & 49.3 & 25.7\\
        & AdaptSegNet\cite{Tsai_adaptseg_2018} & \textbf{69.4} & 14.4 & 48.7 & 0.0 & 0.0 & 11.8 & 2.3 & 23.0 & \textbf{82.4} & 21.7 & 59.0 & 30.3\\
        & Ours (Split) & 65.3 & 22.4 & \textbf{54.6} & 0.2 & 0.0 & \textbf{15.1} & 2.0 & 29.3 & 78.7 & 24.0 & 57.8 & 31.8\\
        & Ours (Fuse) & 65.5 & \textbf{24.7} & 50.0 & \textbf{1.0} & \textbf{0.2} & 12.0 & \textbf{5.3} & 36.7 & 76.2 & \textbf{26.6} & \textbf{60.7} & \textbf{32.6}\\
   \hline
   
   \hline
    \end{tabular}
    }
    \caption{Per-Class IoU on the compound target domain and open domain of the OCDA benchmark: SYNTHIA-SF $\rightarrow$ BDD100K. The results are reported over 11 classes. The best results are denoted in bold. Open $^\dagger$ is open domain covering the BDD100K overcast images.}
    \label{tab:synsf_perclass_opentarget}
\end{table*}

\begin{table*}[!ht]
    \centering
    \resizebox{\textwidth}{!}{
    \begin{tabular}{c|c|cccccccccccccccccc|c}
    \hline
    
    \hline
        \multicolumn{21}{c}{GTA5$\rightarrow$BDD100K}\\
        \hline
        Weather&Method&\rotatebox{90}{road}&\rotatebox{90}{sidewalk}&\rotatebox{90}{building}&\rotatebox{90}{wall}&\rotatebox{90}{fence}&\rotatebox{90}{pole}&\rotatebox{90}{traffic light}&\rotatebox{90}{traffic sign}&\rotatebox{90}{vegetation}&\rotatebox{90}{terrian}&\rotatebox{90}{sky}&\rotatebox{90}{person}&\rotatebox{90}{rider}&\rotatebox{90}{car}&\rotatebox{90}{truck}&\rotatebox{90}{bus}&\rotatebox{90}{train}&\rotatebox{90}{motorbike}&mIoU\\
        \hline
        \multirow{6}{*}{Rainy} & Source\cite{liu2020open} & 48.3 & 3.4 & 39.7 & 0.6 & 12.2 & 10.1 & \textbf{5.6} & 5.1 & 44.3 & 17.4 & 65.4 & 12.1 & 0.4 & 34.5 & 7.2 & 0.1 & 0.0 & 0.5 & 16.2 \\
        & AdaptSegNet\cite{Tsai_adaptseg_2018, liu2020open}& 58.6 & 17.8 & 46.4 & 2.1 & \textbf{19.6} & 15.6 & 5.0 & 7.7 & 55.6 & \textbf{20.7} & 65.9 & 17.3 & 0.0 & 41.3 & 7.4 & 3.1 & 0.0 & 0.0 & 20.2 \\
        & CBST\cite{zou2018unsupervised, liu2020open} & 59.4 & 13.2 & 47.2 & 2.4 & 12.1 & 14.1 & 3.5 & 8.6 & 53.8 & 13.1 & \textbf{80.3} & 13.7 & \textbf{17.2} & \textbf{49.9} & 8.9 & 0.0 & 0.0 & \textbf{6.6} & 21.3 \\
        & IBN-Net\cite{pan2018two, liu2020open} & 58.1 & 19.5 & 51.0 & 4.3 & 16.9 & \textbf{18.8} & 4.6 & \textbf{9.2} & 44.5 & 11.0 & 69.9 & 20.0 & 0.0 & 39.9 & 8.4 & 15.3 & 0.0 & 0.0 & 20.6 \\
        & OCDA\cite{liu2020open} & 63.0 & 15.4 & \textbf{54.2} & 2.5 & 16.1 & 16.0 & \textbf{5.6} & 5.2 & 54.1 & 14.9 & 75.2 & 18.5 & 0.0 & 43.2 & 9.4 & 24.6 & 0.0 & 0.0 & 22.0 \\
        & Ours & \textbf{66.8} & \textbf{22.0} & 52.4 & \textbf{6.7} & 16.7 & 16.9 & 5.3 & 3.5 & \textbf{60.4} & 17.2 & 80.1 & \textbf{21.8} & 0.1 & 46.4 & \textbf{17.9} & \textbf{29.4} & 0.0 & 0.0 & \textbf{24.4}\\
        \hline
         \multirow{6}{*}{Snowy} & Source\cite{liu2020open} & 50.8 & 4.7 & 45.1 & 5.9 & \textbf{24.0} & 8.5 & 10.8 & 8.7 & 35.9 & 9.4 & 60.5 & 17.3 & 0.0 & 47.7 & 9.7 & 3.2 & 0.0 & 0.7 & 18.0\\
        & AdaptSegNet\cite{Tsai_adaptseg_2018, liu2020open}& 59.9 & 13.3 & 52.7 & 3.4 & 15.9 & 14.2 & 12.2 & 7.2 & 51.0 & \textbf{10.8} & 72.3 & 21.9 & 0.0 & 55.0 & 11.3 & 1.7 & 0.0 & 0.0 & 21.2\\
        & CBST\cite{zou2018unsupervised, liu2020open} & 59.6 & 11.8 & 57.2 & 2.5 & 19.3 & 13.3 & 7.0 & \textbf{9.6} & 41.9 & 7.3 & 70.5 & 18.5 & 0.0 & 61.7 & 8.7 & 1.8 & 0.0 & 0.2 & 20.6\\
        & IBN-Net\cite{pan2018two, liu2020open} & 61.3 & 13.5 & 57.6 & 3.3 & 14.8 & \textbf{17.7} & 10.9 & 6.8 & 39.0 & 6.9 & 71.6 & 22.6 & 0.0 & 56.1 & 13.8 & 20.4 & 0.0 & 0.0 & 21.9\\
        & OCDA\cite{liu2020open} & 68.0 & 10.9 & 61.0 & 2.3 & 23.4 & 15.8 & 12.3 & 6.9 & 48.1 & 9.9 & 74.3 & 19.5 & 0.0 & 58.7 & 10.0 & 13.8 & 0.0 & 0.1 & 22.9\\
        & Ours & \textbf{71.8} & \textbf{16.9} & \textbf{61.1} & \textbf{6.5} & 21.4 & 16.3 & \textbf{17.0} & 7.5 & \textbf{52.9} & 8.7 & \textbf{79.7} & \textbf{29.2} & \textbf{0.5} & \textbf{62.7} & \textbf{18.9} & \textbf{29.4} & 0.0 & \textbf{22.6} & \textbf{27.5}\\
        \hline
        \multirow{6}{*}{Cloudy} & Source\cite{liu2020open} & 47.0 & 8.8 & 33.6 & 4.5 & 20.6 & 11.4 & \textbf{13.5} & 8.8 & 55.4 & 25.2 & 78.9 & 20.3 & 0.0 & 53.3 & 10.7 & 4.6 & 0.0 & 0.0 & 20.9\\
        & AdaptSegNet\cite{Tsai_adaptseg_2018, liu2020open}& 51.8 & 15.7 & 46.0 & 5.4 & 25.8 & 18.0 & 12.0 & 6.4 & 64.4 & 26.4 & 82.9 & 24.9 & 0.0 & 58.4 & 10.5 & 4.4 & 0.0 & 0.0 & 23.8\\
        & CBST\cite{zou2018unsupervised, liu2020open} & 56.8 & 21.5 & 45.9 & 5.7 &  19.5 & 17.2 & 10.3 & 8.6 & 62.2 & 24.3 & \textbf{89.4} & 20.0 & 0.0 & 58.0 & 14.6 & 0.1 & 0.0 & 0.1 & 23.9\\
        & IBN-Net\cite{pan2018two, liu2020open} & 60.8 & 18.1 & 50.5 & 8.2 & 25.6 & \textbf{20.4} & 12.0 & \textbf{11.3} & 59.3 & 24.7 & 84.8 & 24.1 & \textbf{12.1} & 59.3 & 13.7 & 9.0 & 0.0 & 1.2 & 26.1\\
        & OCDA\cite{liu2020open} & 69.3 & 20.1 & 55.3 & 7.3 & 24.2 & 18.3 & 12.0 & 7.9 & 64.2 & \textbf{27.4} & 88.2 & 24.7 & 0.0 & 62.8 & 13.6 & 18.2 & 0.0 & 0.0 & 27.0\\
        & Ours & \textbf{79.6} & \textbf{21.7} & \textbf{61.4} & \textbf{11.0} & \textbf{27.6} & 19.4 & 13.4 & 8.3 & \textbf{69.0} & 26.4 & 89.1 & \textbf{25.0} & 3.2 & \textbf{69.5} & \textbf{22.7} & \textbf{21.5} & 0.0 & \textbf{3.5} & \textbf{30.1}\\
        \hline
        \multirow{6}{*}{Overcast} & Source\cite{liu2020open} & 46.6 & 9.5 & 38.5 &  2.7 & 19.8  & 12.9 & \textbf{9.2} & 17.5 &  52.7 &  19.9 & 76.8 &  20.9 & 1.4 & 53.8 & 10.8 & 8.4 & 0.0 & 1.8 & 21.2\\
        & AdaptSegNet\cite{Tsai_adaptseg_2018, liu2020open}& 59.5 & 24.0 & 49.4 &  6.3 & 23.3 & 19.8 & 8.0 & 14.4 & 61.5 & 22.9 & 74.8 & 29.9 & 0.3 & 59.8 &  12.8 & 9.7 & 0.0 & 0.0 & 25.1 \\
        & CBST\cite{zou2018unsupervised, liu2020open} & 58.9 & 26.8 & 51.6 & 6.5 & 17.8 & 17.9 & 5.9 & \textbf{17.9} & 60.9 & 21.7 & \textbf{87.9} & 22.9 & 0.0 & 59.9 & 11.0 & 2.1 & 0.0 & 0.2 & 24.7\\
        & IBN-Net\cite{pan2018two, liu2020open} & 62.9 & 25.3 & 55.5 & 6.5 &  21.2 & \textbf{22.3} & 7.2 & 15.3 & 53.3 & 16.5 & 81.6 & 31.1 & 2.4 & 59.1 & 10.3 & 14.2 & 0.0 & 0.0 & 25.5\\
        & OCDA\cite{liu2020open} & 73.5 & 26.5 & 62.5 & 8.6 & 24.2 & 20.2 & 8.5 & 15.2 & 61.2 & 23.0 & 86.3 & 27.3 & 0.0 & 64.4 & 14.3 & 13.3 & 0.0 & 0.0 & 27.9\\
        & Ours & \textbf{80.1} & \textbf{28.6} & \textbf{66.0} & \textbf{13.0} & \textbf{26.6} & 20.9 & 8.9 & 15.5 & \textbf{67.0} & \textbf{25.1} & 87.7 & \textbf{33.2} & \textbf{9.5} & \textbf{69.2} & \textbf{23.0} & \textbf{18.3} & \textbf{2.2} & \textbf{2.0} & \textbf{31.4}\\
   \hline
   
   \hline
    \end{tabular}
    }
    \caption{Per-Class IoU on different weather images of the OCDA benchmark: GTA5 $\rightarrow$ BDD100K. The rainy, snowy and cloudy weather compose the compound target domain, while the overcast weather is the open domain. The results are reported over 19 classes. The 'bicycle' class is not listed due to the result is close to zero. The best results are denoted in bold.}
    \label{tab:gtaperclass_target}
\end{table*}

\begin{figure*}
    \centering
    \includegraphics[width=\textwidth]{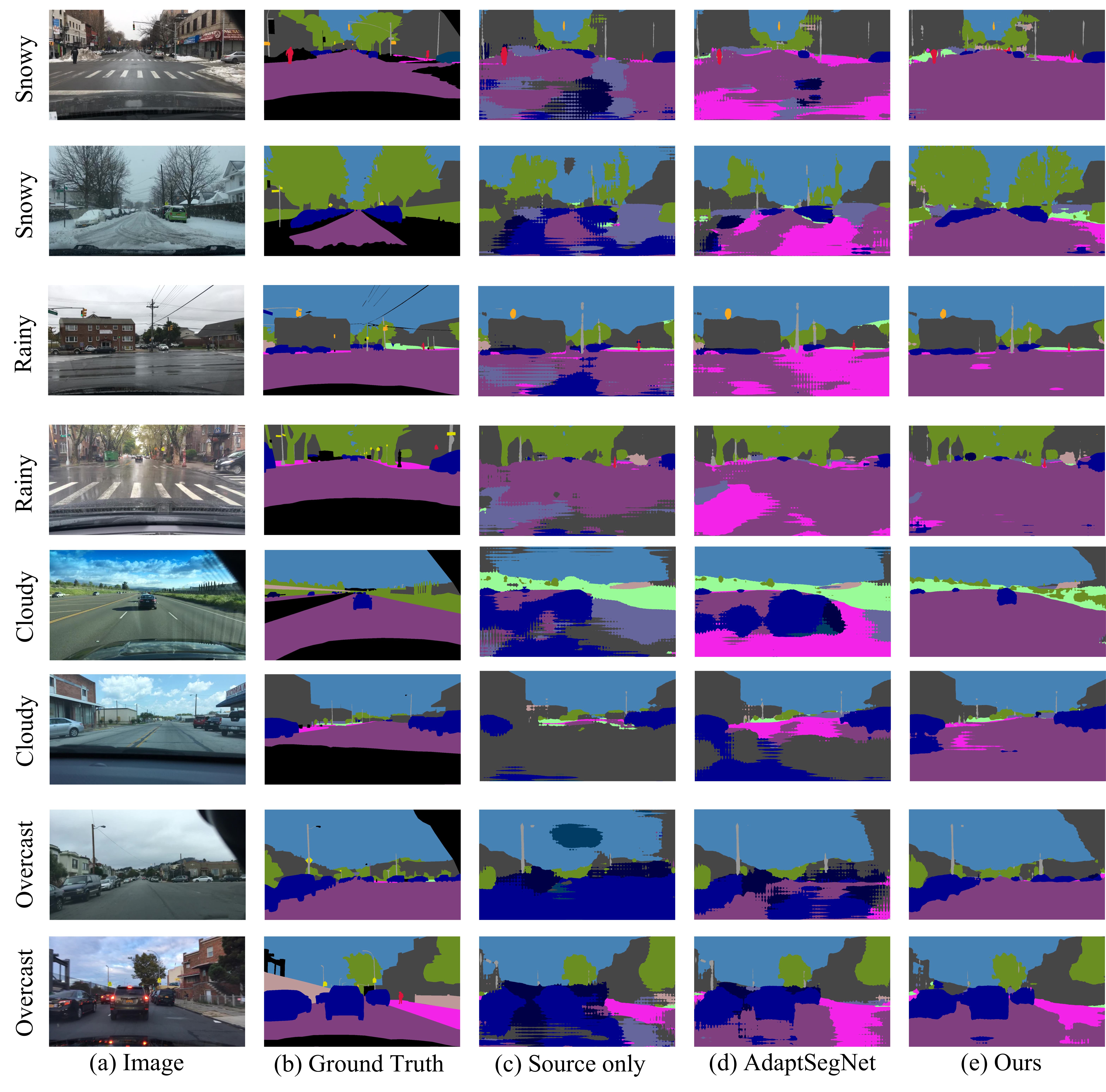}
    \caption{Qualitative semantic segmentation results of the OCDA benchmark: GTA$\rightarrow$ BDD100K. The snowy, rainy and cloudy images are from the compound target domain, while the overcast image is from the open domain. It can be observed that our MOCDA model outperforms the source-only baseline and the AdaptSegNet method on both of the compound target domain and the open domain.}
    \label{fig:seg_vis_supp}
\end{figure*}

\begin{figure*}
    \centering
    \includegraphics[width=\textwidth]{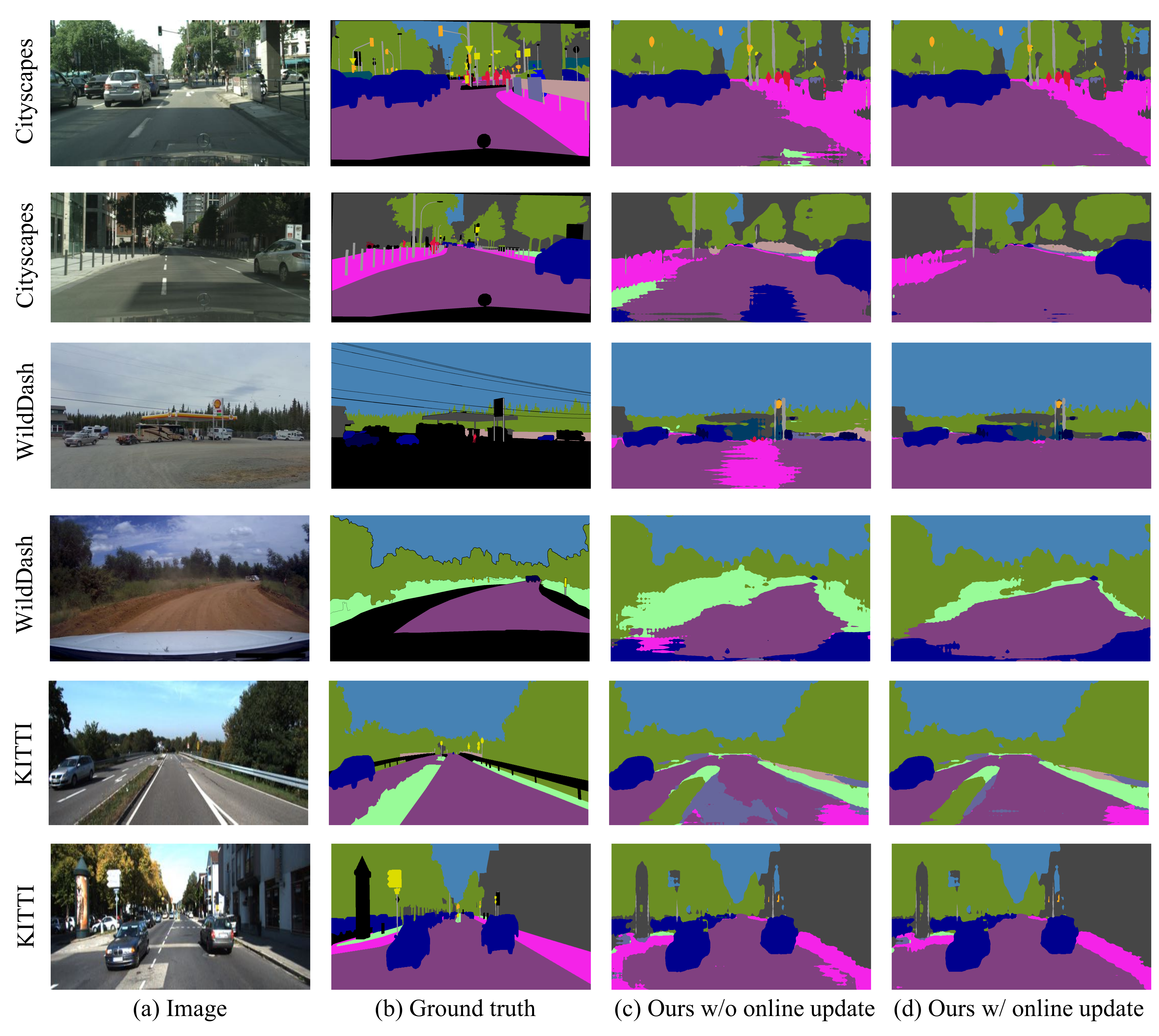}
    \caption{Qualitative semantic segmentation results on the extended open domains of the OCDA benchmark: GTA$\rightarrow$ BDD100K. It is observed that the online update shows obvious benefit for the generalization to the extended open domains.}
    \label{fig:seg_vis_supp_extendedopen}
\end{figure*}

\textbf{Qualitative results.} In Fig. \textcolor{red}{4} of the main paper, we show the qualitative experimental results of our MOCDA model for the OCDA task, on the compound target domain, the open domain and the extended open domains. In Fig.~\ref{fig:seg_vis_supp}, we show more qualitative comparison between our MOCDA model and other methods, on the compound target domain (rainy, snowy and cloudy images), and the open domain (overcast images). It further proves the validity of our MOCDA model for the OCDA task, on both of the compound target domain and the open domain. In Fig.~\ref{fig:seg_vis_supp_extendedopen}, we provide additional qualitative comparison between our MOCDA model with or without online update, on the extended open domains. As shown in Fig.~\ref{fig:seg_vis_supp_extendedopen}, the online update introduces obvious benefit for improving the generalization of the MOCDA model to the extended open domains.

\section{Additional Visualization}

\begin{figure*}
    \centering
    \includegraphics[width=\linewidth]{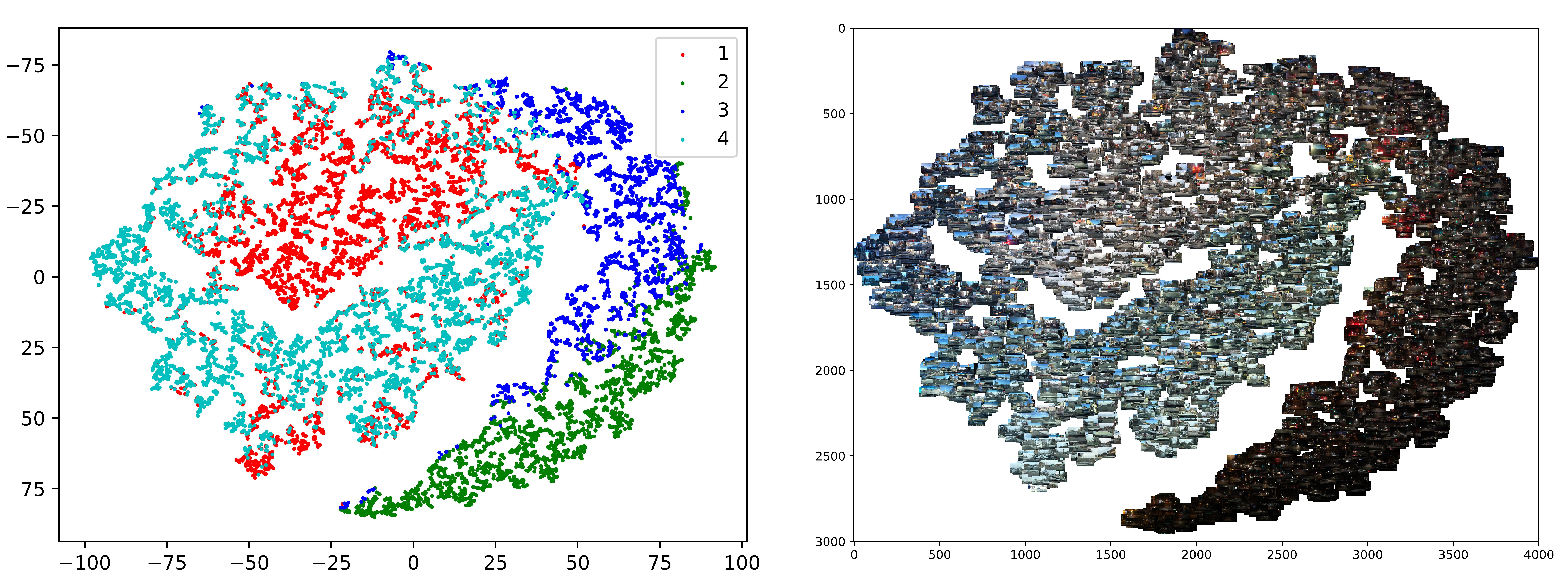}
    \caption{t-SNE visualization of hypernetwork prediction. For image samples belonging to different sub-target domains $1,2,3,4$, our hypernetwork prediction shows different attributes even though we do not explicitly input the sub-target domain information during the fuse module training, which proves the validity of our hypernetwork.}
    \label{fig:hyperweight_visualization}
\end{figure*}

\textbf{Hypernetwork prediction.} In Sec. \textcolor{red}{4.2} of the main paper, we use the ablation study and the variants of our model to prove the validity of the hypernetwork in our MOCDA model. Here we provide additional t-SNE \cite{maaten2008visualizing:tsne} visualization of our hypernetwork prediction. As shown in Fig.~\ref{fig:hyperweight_visualization}, for the image samples from different sub-target domains, our hypernetwork prediction possesses different feature attributes, even though we do not explicitly provide the sub-target domain information in this process. It proves that our hypernetwork is able to adaptively adjust the prediction, conditioned on the style code of the image samples.

\begin{figure*}
    \centering
    \includegraphics[width=\linewidth]{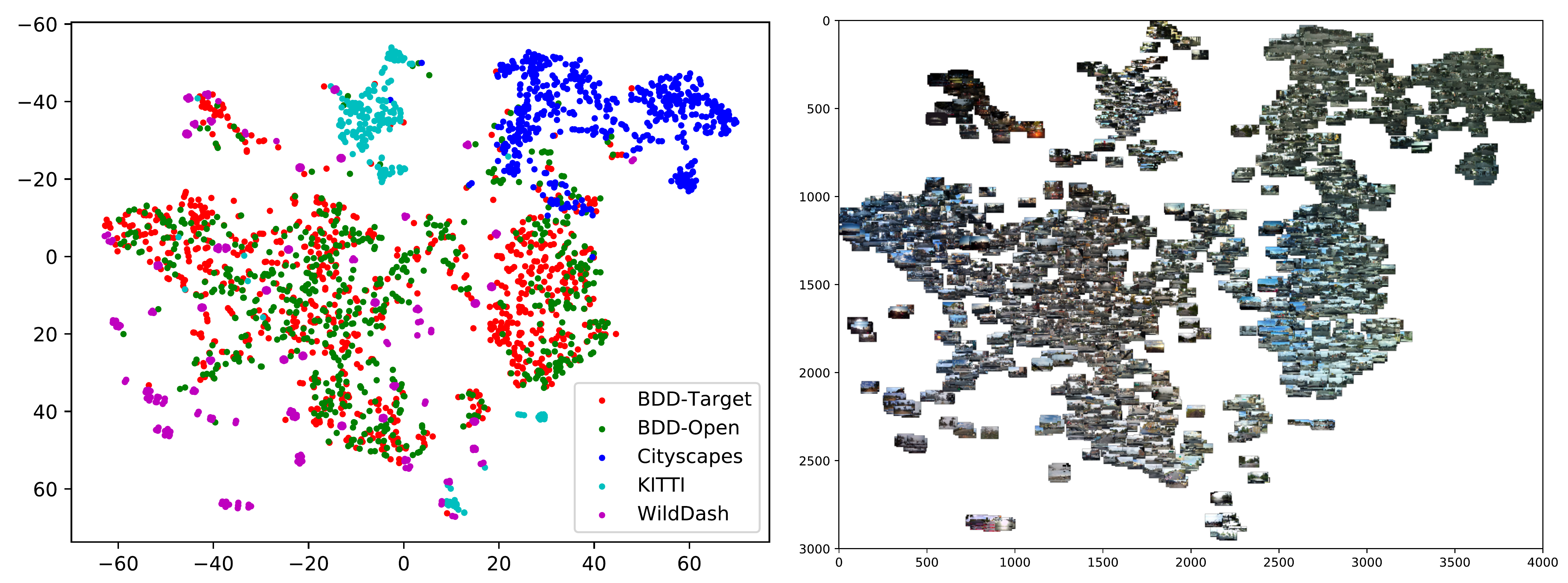}
    \caption{Extended open domains, open domain and target domain style code t-SNE visualization. The domain gap between the BDD100K open domain image and the target domain image (red and green points) is narrow due to the similar style. Our introduced extended open domain Cityscapes, KITTI and WildDash images have much larger domain gap from the BDD100K images. And the style code extracted by our cluster module can effectively reflect the domain gap.}
    \label{fig:open_vis}
\end{figure*}

\textbf{Style code.} In Sec. \textcolor{red}{4} of the main paper, besides the open domain from BDD100K dataset adopted by \cite{liu2020open}, we introduce the extended open domains, which have much larger domain gap to the compound target domain than the open domain from BDD100K dataset, to further measure the generalization ability of the model trained for OCDA task. Here we provide the style code t-SNE \cite{maaten2008visualizing:tsne} visualization of the compound target domain, the open domain and the extended open domains. As shown in Fig. \ref{fig:open_vis}, it can be observed that the domain gap between the open domain and the compound target domain from BDD100K dataset is narrow due to the similar style. Instead, our introduced extended open domains, Cityscapes, KITTI and WildDash dataset, have much larger domain gap from the compound target domain. And the style code extracted by our MOCDA model can effectively reflect the domain gap. It demonstrates the effectiveness of the style code extracted in our MOCDA model, and proves the rationality of our introduced extended open domains for further evaluating the generalization performance of the model to the unseen domains.

